\documentclass[letterpaper]{article}
\pdfoutput=1
\usepackage{uai2020}
\usepackage[margin=1in]{geometry}

\usepackage{times}
\usepackage[numbers]{natbib}
\usepackage[hyphens]{url}  
\usepackage{graphicx}  
\usepackage[vlined,ruled,linesnumbered]{algorithm2e}
\usepackage{subfigure}
\usepackage{color}
\usepackage{amsmath}
\usepackage{booktabs}
\usepackage{enumerate}
\usepackage{amsthm}
\usepackage{amsfonts}

\title{Learning Behaviors with Uncertain Human Feedback}

\author{ {\bf Xu He$^1$, Haipeng Chen$^2$, Bo An$^1$} \\
$^1$ School of Computer Science and Engineering, Nanyang Technological University, Singapore\\
$^2$ Department of Computer Science, Dartmouth College, United States\\
hexu0003@e.ntu.edu.sg, haipeng.chen@dartmouth.edu, boan@ntu.edu.sg\\
}

\begin{document}

\maketitle

\begin{abstract}
Human feedback is widely used to train agents in many domains. However, previous works rarely consider the uncertainty when humans provide feedback, especially in cases that the optimal actions are not obvious to the trainers. For example, the reward of a sub-optimal action can be stochastic and sometimes exceeds that of the optimal action, which is common in games or real-world. Trainers are likely to provide positive feedback to sub-optimal actions, negative feedback to the optimal actions and even do not provide feedback in some confusing situations. Existing works, which utilize the Expectation Maximization (EM) algorithm and treat the feedback model as hidden parameters, do not consider uncertainties in the learning environment and human feedback. To address this challenge, we introduce a novel feedback model that considers the uncertainty of human feedback. However, this incurs intractable calculus in the EM algorithm. To this end, we propose a novel approximate EM algorithm, in which we approximate the expectation step with the Gradient Descent method. Experimental results in both synthetic scenarios and two real-world scenarios with human participants demonstrate the superior performance of our proposed approach.
\end{abstract}

\section{Introduction}
Training agents for tasks with human feedback has many application scenarios such as games and recommendation systems. Previous works \cite{loftin2016learning,loftin2014strategy} recognize the fact that humans as trainers may have different training strategies, i.e., different habits of giving positive/negative feedback, or not giving feedback in some confusing cases. 
They usually solve the problem using Expectation Maximization (EM) by treating human feedback model as hidden parameters. 
However, in environments with uncertainties, trainers may not know exactly the optimal action themselves, and therefore could only provide feedback with uncertainty. 
For example, in many games like Texas hold'em poker, the rewards of all the actions are not deterministic. A sub-optimal action may get huge reward in some cases. Then, trainers are likely to provide positive feedback to the sub-optimal action or not sure which kind of feedback they should give.
Moreover, existing works do not differentiate non-optimal actions, i.e., they assume the expected rewards of different sub-optimal actions are the same, whereas in practice, different actions might have different levels of optimality, and the expected rewards of different sub-optimal actions are usually dependent on their distances to the optimal one.

In this paper, we address these two issues by proposing a novel human feedback model, which takes the uncertainty of trainer's feedback into account. 
Instead of assuming that humans will only give positive feedback when the action taken by the agent is optimal, we model the way that humans give feedback as a probability distribution (e.g., Gaussian), where the probability of giving positive (negative) feedback increases (decreases) w.r.t. the distance of action towards the optimal action. It brings new challenges, in the sense that the newly introduced probabilistic human feedback model makes the calculation of the integration intractable. To this end, we propose a novel approximate EM algorithm, which decomposes the original EM algorithm into two procedures. First, we use an approximated EM step where we assume a subset of the parameters in the human feedback model (e.g., standard deviation of the Gaussian) are fixed and known \textit{a-priori}. In the second step, we use the Gradient Descent (GD) method to compute the most likely parameter values that we fix in the first step.

Our key contributions are summarized as follows.
First, we propose a novel probabilistic human feedback model, which considers the uncertainties in humans while giving feedback to agents in an uncertain environment.
Second, to handle the new computational challenges brought by the new feedback model, we propose a novel approximate EM approach, which integrates Expectation-Maximization (EM) and Gradient Descent (GD).
Last, We conduct extensive experimental evaluations of our proposed approach on both synthetic scenarios and two different real-world scenarios with human participants. 1) We show that our approach outperforms existing works under various settings and metrics. 2) Ablation study shows the effectiveness of our proposed techniques. 3) Robustness analysis demonstrates that our algorithm still performs well even if trainers do not follow our feedback model.

\section{Related Work}
If we only consider the problem in one scenario, it is similar to the multi-armed bandits problem, in which algorithms need to choose a bandit to obtain the maximum reward. However, traditional bandits algorithms like UCB~\cite{auer2002finite} do not consider the human feedback model. 

We now briefly review works in which agents learn form human. The most related domain is interactive reinforcement learning, which aims at hastening the learning process via human feedback. Many existing works \cite{cederborg2015policy,griffith2013policy,knox2009interactively,loftin2014strategy} show the efficiency of learning from human trainers. Knox et al. \cite{Knox2012,knox2009interactively,Knox2010} propose the TAMER framework in which human feedback is learned by a supervised learning method and then used to provide rewards to the agent. The TAMER framework is merged with the deep neural network to address more difficult problems \cite{warnell2017deep}. MacGlashan et al. \cite{Macglashan2017ICML} propose a novel Actor-Critic algorithm considering that human feedback relates to agents' policy. Policy shaping algorithms \cite{cederborg2015policy,griffith2013policy} utilize the difference between `good' and `bad' labels to determine the best action by a Bayesian approach.
How a trainer implicitly expects a learner to interpret feedback is studied in \cite{ho2015teaching}.
Trainers are allowed to directly slightly change the action in paper \cite{perez2018interactive}.
Two works \cite{chevalier2018babyai,macglashan2014training} focus on training agents to learn the meaning of language or commands.

However, most of these approaches do not consider the feedback model of trainers except Loftin et al. \cite{loftin2016learning,loftin2014strategy,macglashan2014training}. They consider trainers' feedback model and propose a strategy-aware algorithm to infer the meaning of no feedback via the trainer's feedback history. However, their model assumes the positive feedback will only be obtained when the agent chooses the optimal action, which is not the case in our problem, as uncertainty affects the feedback provided by trainers. In many cases, trainers generally give positive feedback as long as the action is close to the optimal one especially when they not sure what is the optimal one. For example, if the rewards of actions are not fixed and follow some distributions, a sub-optimal action would obtain more reward than the optimal one. Then, it is hard to only provide positive feedback to the optimal action.

\section{Modelling Uncertain Trainer Feedback}
In one time period, we define state $s \in S$, where $S$ is the set of states. After observing current state $s$, the agent chooses an action $a \in A$. Following~\cite{loftin2016learning}, we define a discrete action space $A=\{a_{1},...,a_{K}\}$.
After the agent chooses an action, the trainer could provide feedback $f \in F =\{f^+,f^-,f^0\}$ to describe his/her degree of satisfaction, where $\{f^+,f^-,f^0\}$ corresponds to positive, negative and no feedback respectively. Positive feedback indicates `good' and negative feedback is `bad', while no feedback could be used to indicate other information such as confusion or ‘not bad’ (but also not good).
All of the trainer's feedback will be recorded in history $h=\{[s_t,a_t,f_t]|t=1,\dots,T\}$, where $T$ is current time period. $s_t\in S$, $a_t\in A$ and $f_t\in F$ are state, action and feedback obtained in the time period $t$.

To model the uncertainty when trainers give feedback, we propose a novel probabilistic human feedback model, the probabilities of obtaining different kinds of feedback are represented by $p(a,\lambda(s);\sigma,\mu,f)$ in state $s$, where $a$ is the current action, $\lambda(s)$ is the trainer's preferred action in state $s$. $\sigma$ and $\mu = \{\mu^+, \ \mu^-\}$ are unknown parameters of the probability function. Formally,
\begin{small}
\begin{align*}
p(a,\lambda(s);\sigma,\mu,f) = \left\{ \begin{array}{ll}
p^+(a,\lambda(s);\sigma,\mu^+) & \textrm{$f=f^+$}\\
p^-(a,\lambda(s);\sigma,\mu^-) & \textrm{$f=f^-$}\\
1-p^+(a,\lambda(s);\sigma,\mu^+)\\
 \quad -p^-(a,\lambda(s);\sigma,\mu^-) & \textrm{$f=f^0$}
\end{array} \right.
\end{align*}
\end{small}
where $\sigma$ and $\mu= \{\mu^+, \ \mu^-\}$ control variance and the maximum probability of providing different kinds of feedback (positive feedback and negative feedback) respectively. The parameters of the model could be adjusted to approximate different trainers' habits of providing feedback. Many popular distributions can be written in this form. For example, the Gaussian function $\mu e^{- \frac{(a-\lambda(s))^2}{2{\sigma}^2}}$ and a simple extension of the density function of the Logistic distribution $\frac{\mu }{\sigma(1+e^{- \frac{(a-\lambda(s))}{\sigma}})}$. 
We can divide the functions $p^+$ and $p^-$ into $\mu \in [0,1]$ and another term $\hat{e}_\sigma \in [0,1]$, formally, $p^+=\mu^+ \hat{e}_\sigma$ and $p^-=\mu^-\left[1-(1-\epsilon)\hat{e}_\sigma\right]$, where $\epsilon$ is a positive constant whose value is close to 0 in order to ensure that the probability of obtaining negative feedback is not 0 at $a=\lambda(s)$.
We constrain that the upper bound of the sum of $p^+(a,\lambda(s);\sigma,\mu^+)$ and $p^-(a,\lambda(s);\sigma,\mu^-)$ cannot be larger than 1 for any $s$ and $a$. Then, we have
\begin{align*}
    &\mu^+\hat{e}_\sigma + \mu^-\left[1-(1-\epsilon)\hat{e}_\sigma\right] \leq 1 \\
    \Rightarrow & 1- \mu^- - \hat{e}_\sigma \left[\mu^+ + \mu^-(1-\epsilon)\right] \geq 0. 
\end{align*}
where $\hat{e}_\sigma \in [0,1]$. Notice that the function on the left side is monotonous for the term $\hat{e}_\sigma$, which means that the minimum of the function is obtained when $\hat{e}_\sigma =1 $ since $\mu \geq 0$ and $\epsilon \rightarrow 0^+$. Thus, we have
\begin{align*}
    1 - \mu^+ - \epsilon \mu^- \geq 0 & \Rightarrow \mu^+ + \epsilon \mu^- \leq 1.
\end{align*}
By using the property $\mu^- \in [0,1]$, we get the upper bound of $\mu^+$:
$$\mu^+ + \epsilon \mu^- \leq 1 \Rightarrow \mu^+ \leq 1-\epsilon.$$


The intuitive interpretation of $\mu^+$ and $\mu^-$ is that they are the maximum probabilities of obtaining positive and negative feedback respectively. That is, $\mu^+$ is the probability of getting positive feedback when the current action is the trainer's most preferred one and $\mu^-$ is the probability of getting negative feedback at the trainer's most disliked action. When the action $a$ is different from $\lambda(s)$, the probability of obtaining positive feedback declines while more negative feedback will be received.

\section{Approximate Expectation Maximization}
Since Bayesian approaches have shown the efficiency in inferring from a small quantity of data, we adopt the Maximum Likelihood Estimation (MLE) to estimate the policy function $\lambda$ with two unknown parameters $\mu$ and $\sigma$ in the probabilistic human feedback model, where the objective function is represented as:
\begin{equation}
\arg\max\nolimits_{\lambda} P(h|\lambda;\mu,\sigma),
\end{equation}
where $h=\{[s_t,a_t,f_t]|t=1,\dots,T\}$ is the history of state, action and feedback.
Note that this MLE cannot be calculated directly since there are two sets of unknown variables, policy $\lambda$ and parameters ($\mu$,$\sigma$), in this equation and there is no explicit bound for $\sigma$.
One classical way of handling this kind of problems is to use the EM algorithm where the $i$-th EM update step is represented as: 
\begin{equation*}
\begin{split}
\lambda_{i+1}=\arg \max\nolimits_\lambda &\iiint P(\mu^+,\mu^-|h,\lambda_i)\\
&\cdot \ln [P(h,\mu^+,\mu^-|\lambda)] d\mu^+ d\mu^- d \sigma.
\end{split}
\end{equation*}
However, it is easy to see that this integral is intractable, as 1) there is no explicit upper bound for $\sigma$ and 2) the large dimension of the hidden variables significantly increases the computational complexity of calculating the integral (which is usually calculated using numerical methods as there are no analytical solutions). As a result, we propose a novel approximate EM algorithm, in which an alternating optimization method is used to solve this MLE by two steps:
1) Update $\lambda$ with $\sigma$ fixed and treat $\mu$ as a hidden variable by EM algorithm. 2) Update $\sigma$ with $\lambda$ fixed by GD algorithm and use a trick to ignore $\mu$. \footnote{An alternative is to use gradient-based methods at both steps \cite{birnbaum1968some}. However, this is infeasible due to the complex problem structure and the difficulty in deriving gradients at both steps.}


\subsection{Updating $\lambda$ with $\sigma$ Fixed}
Since $\mu^+$ and $\mu^-$ mean the maximum probabilities of obtaining positive and negative feedback, we know that the range of $\mu^-$ is $[0,1]$ and $0 \leq \mu^+ \leq 1 - \epsilon$. By treating $\mu^+$ and $\mu^-$ as bounded latent variables, $\lambda$ can be updated by the Expectation-Maximization (EM) algorithm when $\sigma$ is fixed, which is inspired by \cite{loftin2014strategy}.
In the expectation step, we compute the expectation of a log-likelihood estimate with respect to a bounded latent variable. In the maximization step, the expectation is maximized with respect to another unknown variable.
Treating $\mu^+,\mu^-$ as latent variables, the $i$-th EM update step is:
\begin{equation}
\begin{split}
\lambda_{i+1}=\arg \max\nolimits_\lambda &\int_0^1 \int_0^{1-\epsilon} P(\mu^+,\mu^-|h,\lambda_i)\\
&\cdot \ln [P(h,\mu^+,\mu^-|\lambda)] d\mu^+ d\mu^-
\end{split},
\end{equation}
where $\lambda_i$ is the inferred preference in $i$-th step. To ensure that the probability of receiving positive and negative feedback is not larger than 1, the upper bound of $\mu^+$ is defined by $1-\mu^-$. 

Using the Bayes' theorem and the property of logarithm, we obtain
\begin{equation} \label{term1}
    P(\mu^+,\mu^-|h,\lambda_i) = \frac{P(h|\mu^+,\mu^-,\lambda_i)P(\mu^+,\mu^-|\lambda_i)}{P(h|\lambda_i)},
\end{equation}
and
\begin{equation} \label{term2}
    \begin{split}
    \ln [P(h,\mu^+,\mu^-|\lambda)] =& \ln [P(h|\mu^+,\mu^-,\lambda)] \\
    &+ \ln [P(\mu^+,\mu^-|\lambda)].
    \end{split}
\end{equation}
Notice that $P(h|\lambda_i)$ is the marginal probability and does not involve any variable. Thus we can treat it as constant and remove it from Eq.(\ref{term1}). Since $\mu^+$ and $\mu^-$ define the way that a trainer provides feedback and $\lambda$ is the set including the optimal actions in various states, $\{\mu^+,\mu^-\}$ is independent of $\lambda$ and thus $P(\mu^+,\mu^-|\lambda) = P(\mu^+,\mu^-|\lambda_i) = P(\mu^+,\mu^-)$.
We assume that $\mu^+$ and $\mu^-$ are uniformly distributed due to the lack of the prior knowledge about $\mu^+$ and $\mu^-$. Therefore, we get $p(\mu^+,\mu^-)=2$ by solving the equation $$\int_0^1 \int_0^{1-\epsilon} p(\mu^+,\mu^-) d\mu^+ d\mu^- = 1.$$
Thus, the term $P(\mu^+,\mu^-|\lambda_i)$ can be removed from Eq.(\ref{term1}). For the second logarithmic term of Eq.(\ref{term2}), since
\begin{equation}
\int_0^1 \int_0^{1-\epsilon} P(\mu^+,\mu^-|h,\lambda_i) \ln [2] d\mu^+ d\mu^-
\end{equation}
is not related to $\lambda$, we can ignore it.
Finally, the objective could be simplified to
\begin{align}
\begin{split}
\lambda_{i+1}(s)=&\arg \max_\lambda \int_0^1 \int_0^{1-\epsilon}P(h|\mu^+,\mu^-,\lambda_i)\\
&\cdot \ln [P(h^s|\mu^+,\mu^-,\lambda(s))] d\mu^+ d\mu^-
\end{split},
\label{EM}
\end{align}
where $h^s$ is the history containing the state $s$.
Utilizing $h=\{[s_t,a_t,f_t]|t=1,\dots,T\}$, we can calculate the integral, since the probability of obtaining $h$ given $\mu$ and $\lambda$ is:
\begin{align*}
P(h|\mu^+,\mu^-,\lambda)=&\prod_{T} P(f_t|a_t,s_t,\lambda(s_t),\mu^+,\mu^-)\\
=&\prod_{T} p(a_t,\lambda(s_t);\sigma,\mu,f_t)\\
=& \prod_{s_h\in S} p^+(a_h,\lambda(s_h);\sigma^+,\mu^+)^{|n_{s_h}^+|} \\
&\cdot p^-(a_h,\lambda(s_h);\sigma^+,\mu^+)^{|n_{s_h}^-|} \\
&\cdot p^0(a_h,\lambda(s_h);\sigma^+,\mu^+)^{|n_{s_h}^0|},
\end{align*}
where $|n_{s_h}^+|,\ |n_{s_h}^-|,\ |n_{s_h}^0|$ are the numbers of three types of feedback in the sate $s_h \in h$.
For $p^+$ and $p^-$, notice that $\ln [\mu \hat{e}_\sigma] = \ln[\mu]+\ln[\hat{e}_\sigma]$. Since $\ln[\mu]$ is not related to $\lambda$, we ignore it during the calculation to prevent divergence.
In practice, we compute expectations for all the actions that are available in a sate $s$ and select the action with the maximal expectation as the policy $\lambda(s)$ for the state $s$.

Eq.(\ref{EM}) shows the natural way to utilize feedback from other states. When we compute the optimal action in state $s$, the first term $P(h|\mu^+,\mu^-,\lambda_i)$ considers all the historical feedback and effects the result of the Eq.(\ref{EM}). 
In this way, both feedback model and historical feedback are taken into consideration to infer the best action.
\IncMargin{1em}
\begin{algorithm}[t]
\SetKwInOut{Initialize}{Initialize}
\SetKwInOut{Input}{Input}
\SetKwProg{Fn}{Function}{}{}
\caption{Adaptive Bayesian Learning with Uncertain Feedback (ABLUF)}\label{ABLUF}
    \Initialize{$\lambda=randomPolicy(),\epsilon=0.01,$\\
    $h=[\ ],t=0,done = 0$\;}
    \BlankLine
    \While{$done \neq 1$}{
        $s_t=getState()$\; \label{begin}
        $a_t=\lambda(s)$\; \label{act}
        $takeAction(a_t)$\; \label{takeact}
        $f_t=getFeedback()$\;
        $h=[h;[s_t,a_t,f_t]]$\;
        \Repeat{$\lambda'=\lambda$}{ \label{em1}
            $\lambda'=\lambda$\;
            $\lambda=\arg \max_\lambda \int_0^1 \int_0^{1-\epsilon}P(h|\mu^+,\mu^-,\lambda')\cdot \ln [P(h^s|\mu^+,\mu^-,\lambda(s))] d\mu^+ d\mu^-$\;
        }\label{em2}
        $\sigma = \sigma - \alpha \frac{1}{n} \sum_s \sum_a \nabla_{\sigma} L(\sigma;f,\lambda)$\; \label{update}
        $done = getDone()$\;
        $t = t+1$\;
}
\end{algorithm}
\DecMargin{1em}
\subsection{Updating $\sigma$ with $\lambda$ Fixed}
Since our objective function is calculated based on probabilities provided by the feedback model, the accuracy of the feedback model affects the result significantly.
To obtain an accurate human feedback model, we use GD method to minimize the square loss function between the inferred feedback model $p(a,\lambda(s);\sigma,\mu,f)$ and the real one, i.e.,
\begin{equation}
[p(a,\lambda(s);\sigma,\mu,f)- p(f|a,s)]^2.
\end{equation}

However, since we calculate the integral under $\mu$, the value of $\mu$ is unknown. We cannot compute the gradient of the loss function. Note the form of function $p(a,\lambda(s);\sigma,\mu,f)$, the exponential term actually is the estimation of the ratio between the probability at action $a$ and the maximum probability of receiving positive or negative feedback. That is,
\begin{equation}
\begin{split}
ratio_p^+=&\hat{e}_\sigma \\
\approx& \frac{p(f^+|a,s)}{p(f^+|\lambda(s),s)}=ratio_a(f^+) \\
ratio_p^-=&[1-(1-\epsilon)\hat{e}_\sigma]\\ 
\approx& \frac{p(f^-|a,s)}{p(f^-|a^-_s,s)} = ratio_a(f^-),
\end{split}
\end{equation}
where $a^-_s=\arg\max_a d(a,\lambda(s))$ is the action that a trainer dislikes the most, since this action has the maximum probability of receiving negative feedback. We transform the loss function to
\begin{equation}
L(\sigma;f,\lambda)= [ratio_p - ratio_a(f)]^2.
\end{equation}
We omit $\mu$ in $L$, since it does not appear in the new formulation of the loss function. 

The gradient $\nabla_{\sigma} L(\sigma;f,\lambda)$ of the new loss function could be computed as follows. For $f^+$, we have
\begin{equation}\label{grad_po}
[ratio_p^+-ratio_a(f^+)]\nabla_{\sigma}ratio_p^+.
\end{equation}
For $f^-$, we have
\begin{equation}\label{grad_ne}
[ratio_p^- -ratio_a(f^-)]\nabla_{\sigma}ratio_p^-.
\end{equation}
Thus,
\begin{equation*}\label{grad}
\begin{split}
    \nabla_{\sigma} L(\sigma;f,\lambda) =& [ratio_p^+-ratio_a(f^+)]\nabla_{\sigma}ratio_p^+ \\
    &+ [ratio_p^- -ratio_a(f^-)]\nabla_{\sigma}ratio_p^-
\end{split},
\end{equation*}
where $ratio_a(f^+)$ and $ratio_a(f^-)$ could be obtained from the historical feedback.
For example, if the probabilities of offering positive feedback are 0.5 and 0.9 in action $a$ and the optimal action $\lambda_i(s)$ respectively, $ratio_a(f^+) = \frac{0.5}{0.9}$. We implement the Gradient Descent method to update the parameters and all the historical feedback of all actions is used to compute the gradient and then update $\sigma$:
\begin{equation}
\label{gd}
\sigma = \sigma - \alpha \frac{1}{n} \sum\nolimits_s \sum\nolimits_a \nabla_{\sigma} L(\sigma;f,\lambda),
\end{equation}
where $n=|S|\times|A|$ is the total number of states-action pairs.

The Adaptive Bayesian Learning with Uncertain Feedback (ABLUF) algorithm is shown in Algorithm \ref{ABLUF}. The value of step size $\alpha$ is given in the experiment section based on the property of the gradient.
The variable $done$ is obtained from the environment and the trainer, which is also introduced in the next section. 
After selecting and taking the action in Lines \ref{act}-\ref{takeact}, the system will obtain the trainer's feedback and record it in $h$. 
In Lines \ref{em1}-\ref{em2}, the EM algorithm calculates the preferences of the trainer by Eq.(\ref{EM}) based on the records stored in $h$. After updating $\lambda$, the parameter $\sigma$ is updated at Line \ref{update} by Eq.(\ref{gd}). 

\section{Experiment Evaluations}

We implement two sets of experiments for two environments with human subjects: training a virtual dog to catch rats and learning users' preferences of lighting.
We also evaluate the performance, convergence, and robustness of our proposed algorithm in simulated experiments. Code can be found at \url{https://github.com/hlhllh/Learning-Behaviors-with-Uncertain-Human-Feedback}.

\subsection{Choice of human feedback model $p^+$ and $p^-$}
In general, a good human feedback model is expected to have the following good properties: 
1) It captures the uncertainty of a trainer to give feedback. 
2) The optimal action is unique in each state.  
3) When the action becomes far away from the optimal by a same amount, trainer's satisfaction will decrease by a similar amount.

For these concerns, we use Gaussian functions as the human feedback model due to its simplicity. \footnote{Our robustness analysis shows that the Gaussian functions model well adapts to situations where the way trainers give feedback is different from the Gaussian functions. Other forms of human feedback model could also be considered, such as the probability density functions of Cauchy distribution, Logistic distribution, etc. We leave the exploration of various forms of human feedback model as future work.} Thus, we define 
\begin{align}
p^+(a,\lambda(s);\sigma,\mu^+) &= \mu^+ e^{- \frac{(a-\lambda(s))^2}{2{\sigma}^2}}\\
p^-(a,\lambda(s);\sigma,\mu^-) &= \mu^-[1-(1-\epsilon) e^{- \frac{(a-\lambda(s))^2}{2{\sigma}^2}}]
\end{align}
and the gradients of $L(\sigma;f,\lambda)$ are $$2\frac{(a-\lambda_i(s))^2}{{\sigma}^3}[ratio_p^+-ratio_a(f^+)]ratio_p^+$$
and 
$$-2(1-\epsilon)\frac{(a-\lambda_i(s))^2}{{\sigma}^3}[ratio_p^- -ratio_a(f^-)]ratio_p^-$$
for $f^+$ and $f^-$ respectively.
The learning rate $\alpha$ is set to be $0.4\sigma^3$. The $\sigma^3$ term aims at eliminating the denominator to avoid a very small update of the parameter $\sigma$ when the value of $\sigma$ is large.

\subsection{Baselines}
We compare with two state-of-the-art algorithms
ISABL \cite{loftin2016learning} is a Bayesian learning algorithm aiming at utilizing human feedback to train an agent. Expectation-Maximization method is used to calculate the best action. 
They assume that the human trainer only provides positive feedback when the agent selects an optimal action. If the agent chooses other actions, the trainer will give negative feedback. No feedback is also considered to determine the training policy. The error rate of the ISABL algorithm in our experiment is 0.1 following the original setting.

The upper confidence bound (UCB) algorithm \cite{auer2002finite} calculates the upper confidence bound of the expected reward for each action and chooses an action with maximum UCB value. In our experiment, we assign values for different kinds of feedback, that is, $[f^+,f^-,f^0] \rightarrow [1,-1,0]$. After receiving the feedback, the upper confidence bound for an action will be computed by
$UCB(s,a)=\mathbb{E}[r_{s,a}]+ \sqrt{\frac{2\log{t_{s,a}}}{t_{s}}},$
where the $\mathbb{E}[r_{s,a}]$ is the expected feedback value conditioning on the feedback for action $a$ in state $s$, $t_{s,a}$ means the number of times that the algorithm chooses $a$ in state $s$, and $t_{s}$ is the total number of times that state $s$ appears.

\subsection{Training a virtual dog}
\begin{figure}[!t]
\setlength{\belowcaptionskip}{-0.2cm}
\centering
\subfigure[State]{\label{chasing_state}
\includegraphics[width=0.22\textwidth]{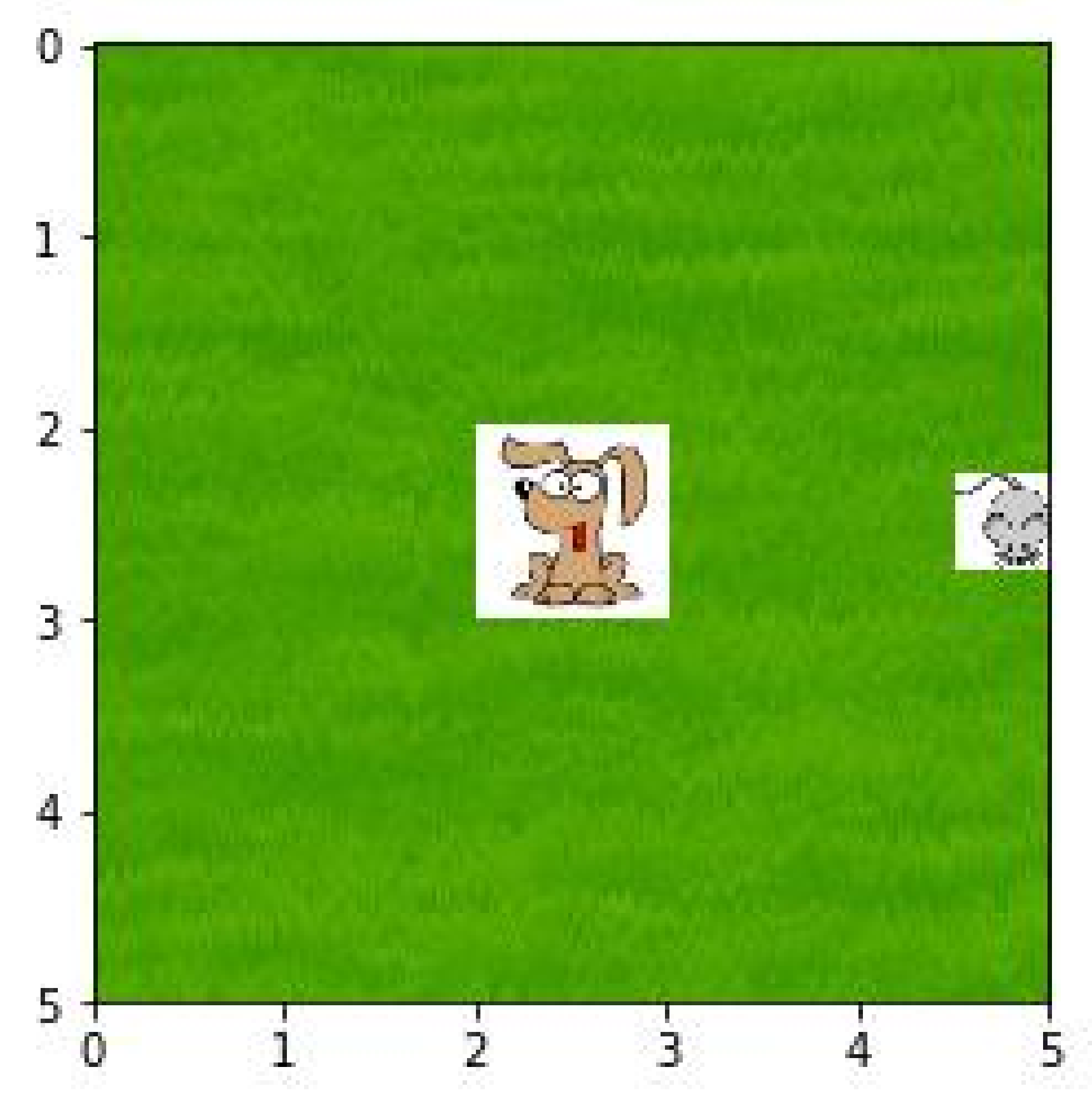}
}
\subfigure[Action]{\label{chasing_action}
\includegraphics[width=0.22\textwidth]{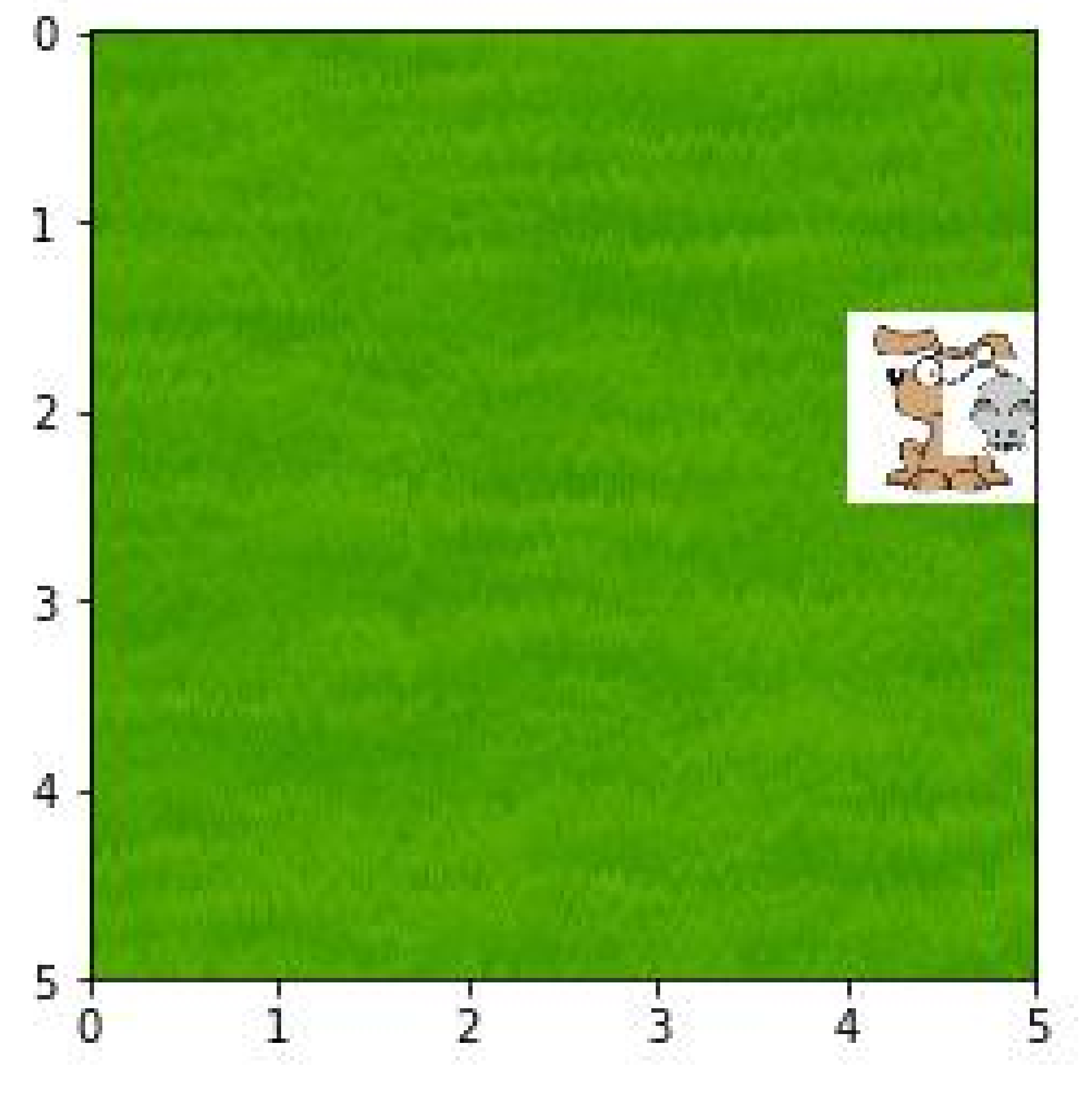}
}
\setlength{\abovecaptionskip}{0pt}
\caption{The state and action of training a dog.}\label{chasing}
\end{figure}

\begin{figure}[!t]
\setlength{\belowcaptionskip}{-0.5cm}
\centering
\subfigure[Step]{\label{human_dog_t}
\includegraphics[width=0.14\textwidth]{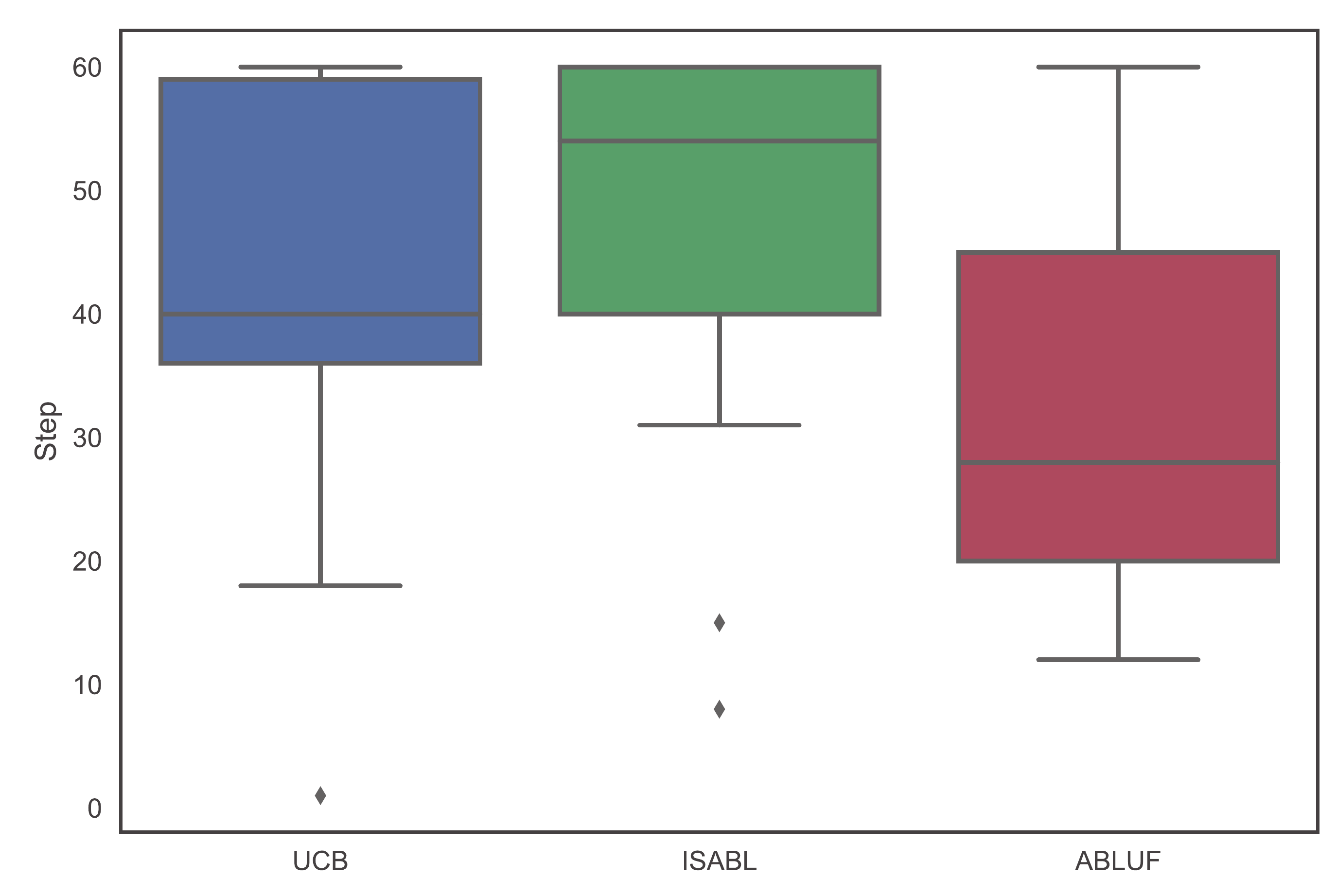}
}
\subfigure[Rats caught per step]{\label{human_dog_p}
\includegraphics[width=0.14\textwidth]{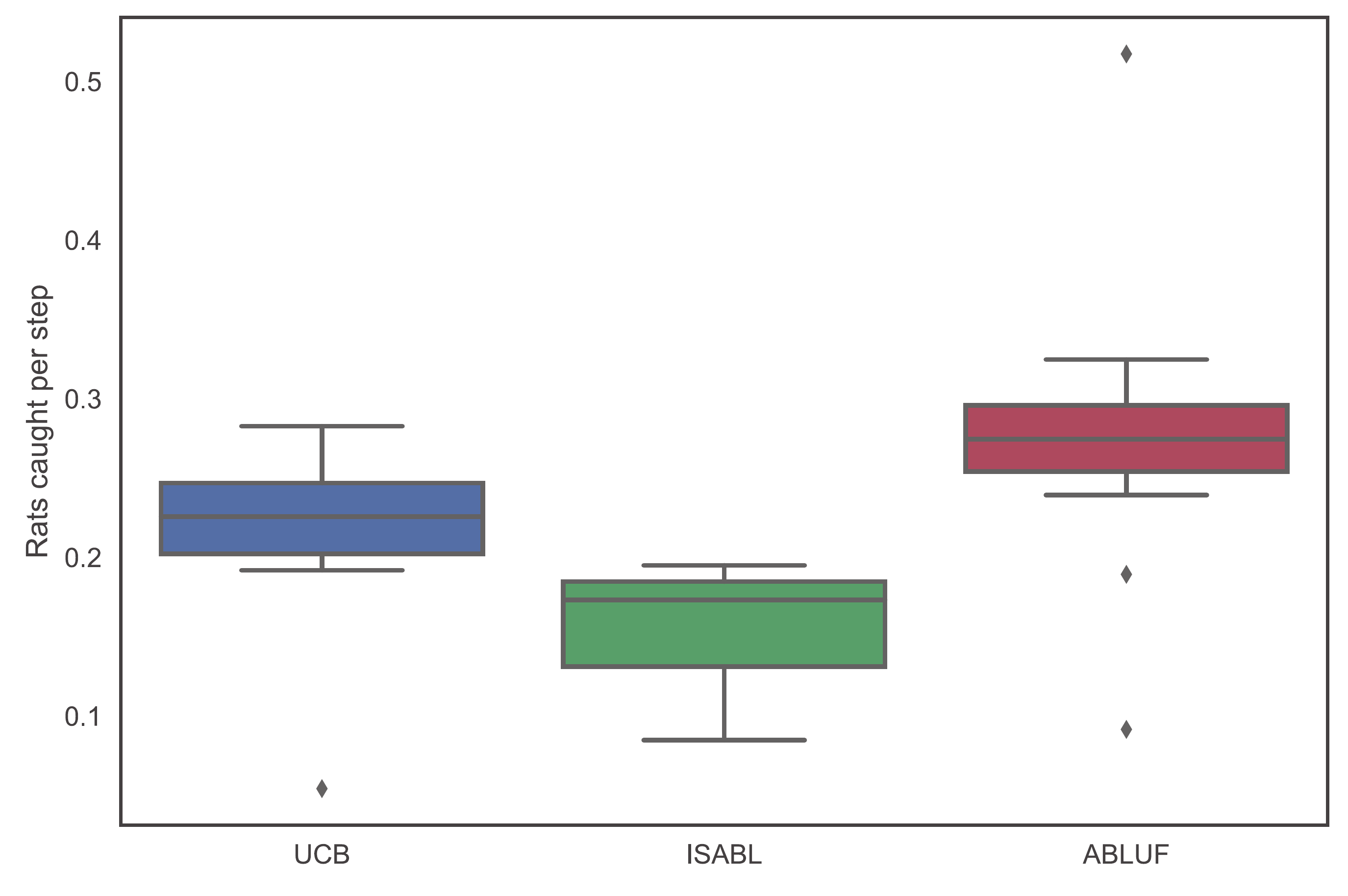}
}
\subfigure[The gap between the learned policy and the optimal policy]{\label{human_dog_distance}
\includegraphics[width=0.14\textwidth]{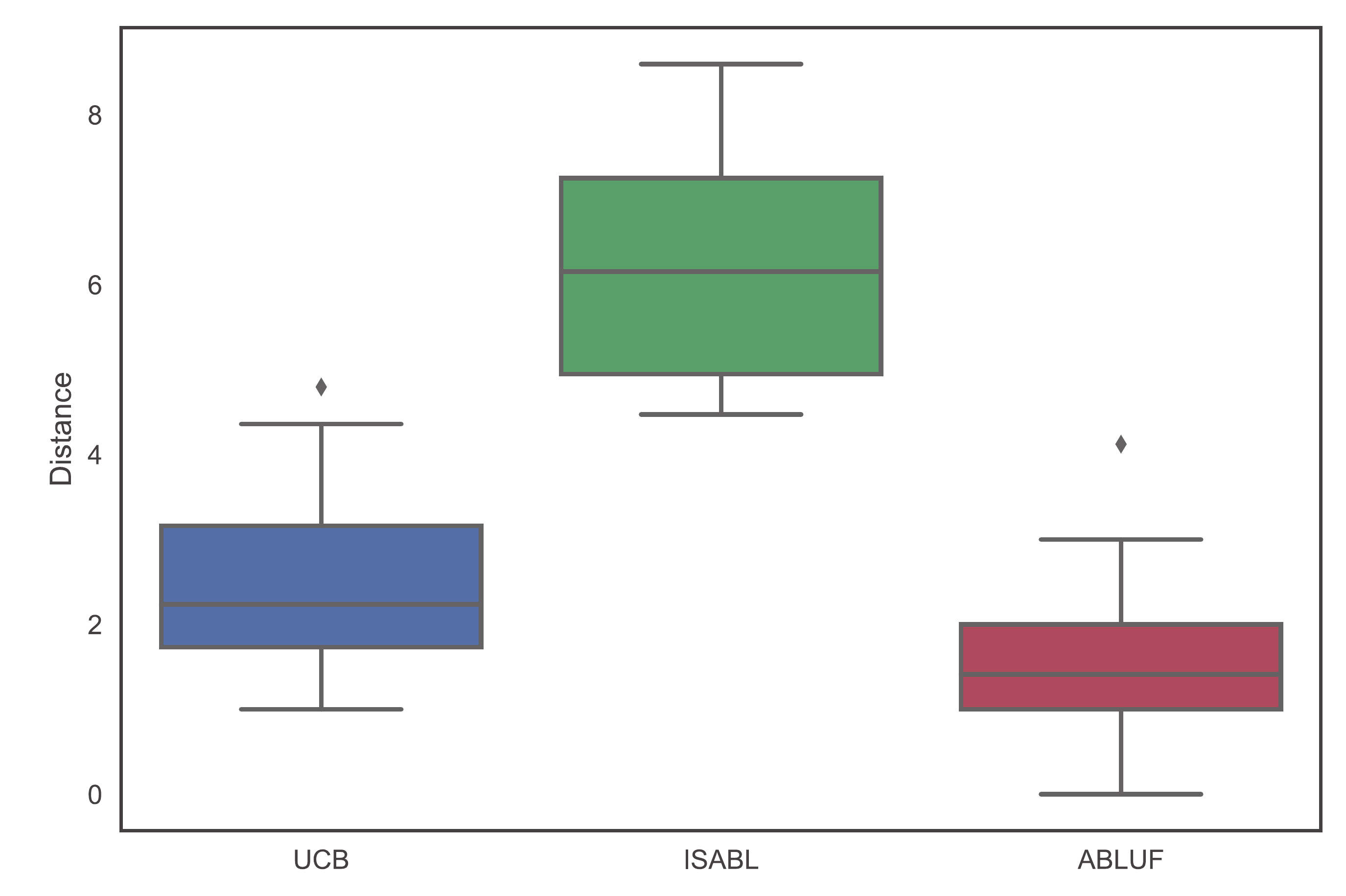}
}
\setlength{\abovecaptionskip}{0pt}
\caption{The performance of different algorithms to train a dog.}\label{human_dog}
\end{figure}
In our first experiment, trainers are required to train a virtual dog to catches rats. At the beginning of each step, the dog is at the center of the figure and a rat appears at one of four edges of the figure. Each edge can be considered as a state in this experiment. Then, the rat moves to one of 6 points on the same edge following a distribution that is fixed for each edge and not influenced by the dog. The dog chooses one point from 6 candidate points to go as the action in this state. The probability that the dog catches the rat is inverse to the distance between the dog and the rat. 
After observing the locations of the dog and the rat, trainers provide feedback to current action of the dog. Then, the next step begins. For example, Figure \ref{chasing_state} is a state and Figure \ref{chasing_action} shows the action of dog and the location the rat appears. 
The optimal actions for all states are generated randomly before the experiment begins. For each state, we use a standard Gaussian function per state to decide the probability that the rat appears at different points, which is $e^{- \frac{(a-\lambda(s))^2}{2}}, \ \forall a$. 
Normalization is executed to ensure the sum of probabilities is 1. Similarly, the probability that the rat is caught is $e^{- \frac{(a_{rat}-a_{dog})^2}{2}}$.
We invited 40 trainers to participate in this experiment. Four states are randomly ordered and appears one by one. 

\textbf{When to stop?} We limit the number of steps that trainers can interact with the agent is 15 in each state in order to constrain the time of the experiment. We will show that this limitation is reasonable since the probability that our algorithm learns the optimal actions under this constraint is high. The state changes when 1) the trainer thinks that the dog's performance is good enough or 2) the number of steps exceeds 15. The experiment ends if all the states are satisfied the above mentioned conditions. Our algorithm and two baselines introduced in the last section are executed one by one with random order for each trainer. Trainers are not informed of the order.

\textbf{Metrics.} Three metrics are applied to judge the performance of various algorithms: 1) The number of steps used to finish the experiment. 2) The average number of caught rats over steps. 3) The 2-norm between the learned policy and the optimal policy. A good algorithm can catch more rats during training and finally learn the optimal policy with less steps.

\textbf{Results.} Results under these metrics are shown in Figure \ref{human_dog}. Figure \ref{human_dog_t} shows that most of trainers finish the training before 15 steps per state. However, for other baselines, more steps are required to satisfy trainers' criteria of ending, which indicates that other algorithms perform relatively badly after training. Figure \ref{human_dog_p} shows the number of rats caught per step. The larger the value is, the faster an algorithm converges to the optimal policy during the training. The gap between the optimal policy and the learned policy is illustrated in Figure \ref{human_dog_distance}. The gap is calculated by the 2-norm between these two policies. These three figures indicate that our algorithm performs the best considering the speed of convergence and the quality of the learned policy. For all the three metrics, the ABLUF method is statistically superior to others using Wilcoxon two-sided signed-rank test ($p < 0.01$).

\subsection{Learning users' preference of lighting}
The second experiment aims at learning users' preference of lighting in different situations. Users provide different kinds of feedback after an action corresponding to a light level is chosen by an algorithm. Since users' utility functions are very difficult for users to access and human sense is inaccurate~\cite{boutilier2002pomdp,knill2004bayesian}, human feedback is uncertain and they are likely to provide positive feedback in some sub-optimal action especially when the difference of two nearby light levels is tiny. To verify this assumption, we introduce the QUERY method as a baseline, which asks users to directly choose light levels.
\begin{figure}[!t]
\centering
\subfigure[QUERY method.]{
\includegraphics[width=0.26\textwidth]{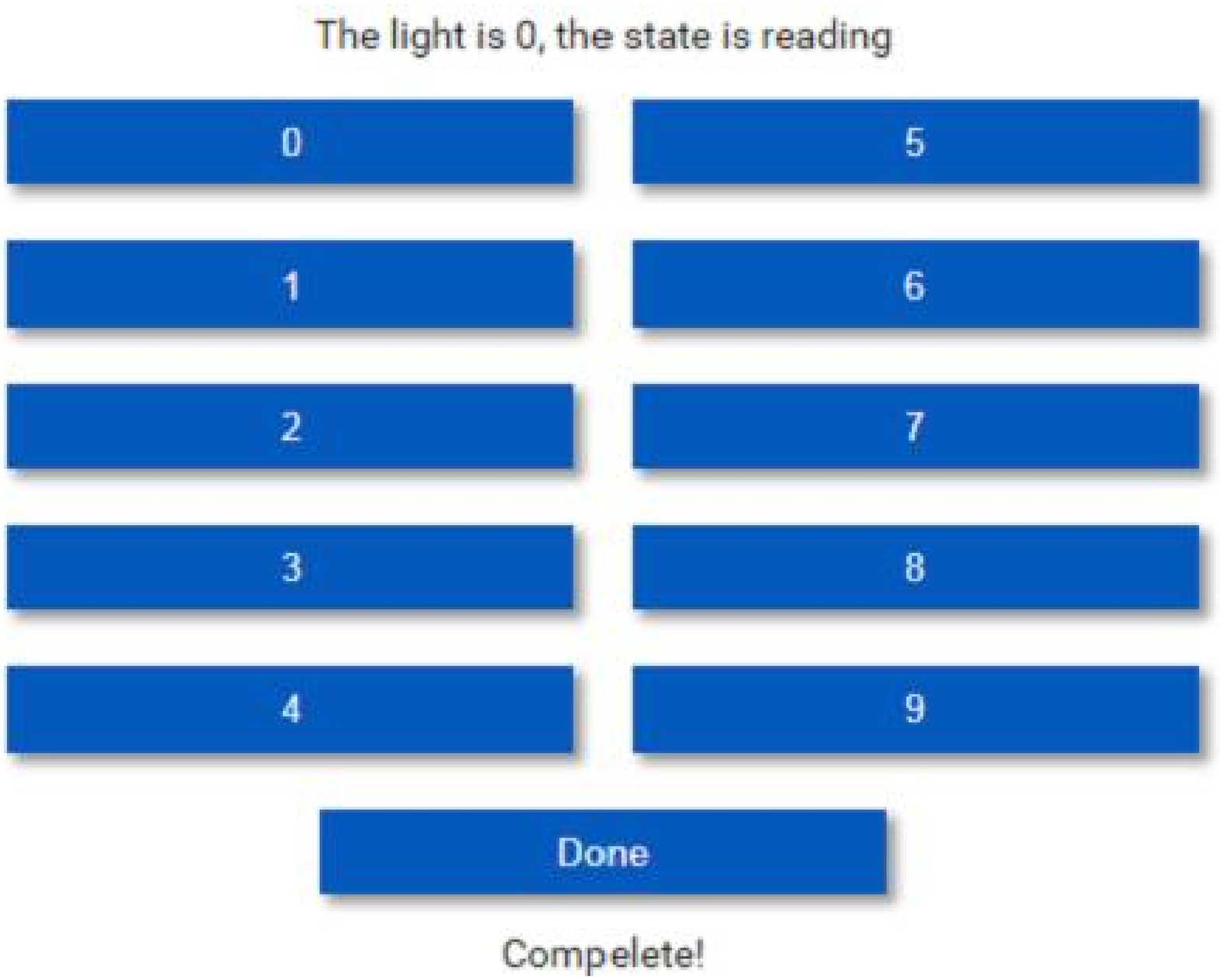}\label{gui_query}
}
\subfigure[Other methods.]{
\includegraphics[width=0.18\textwidth]{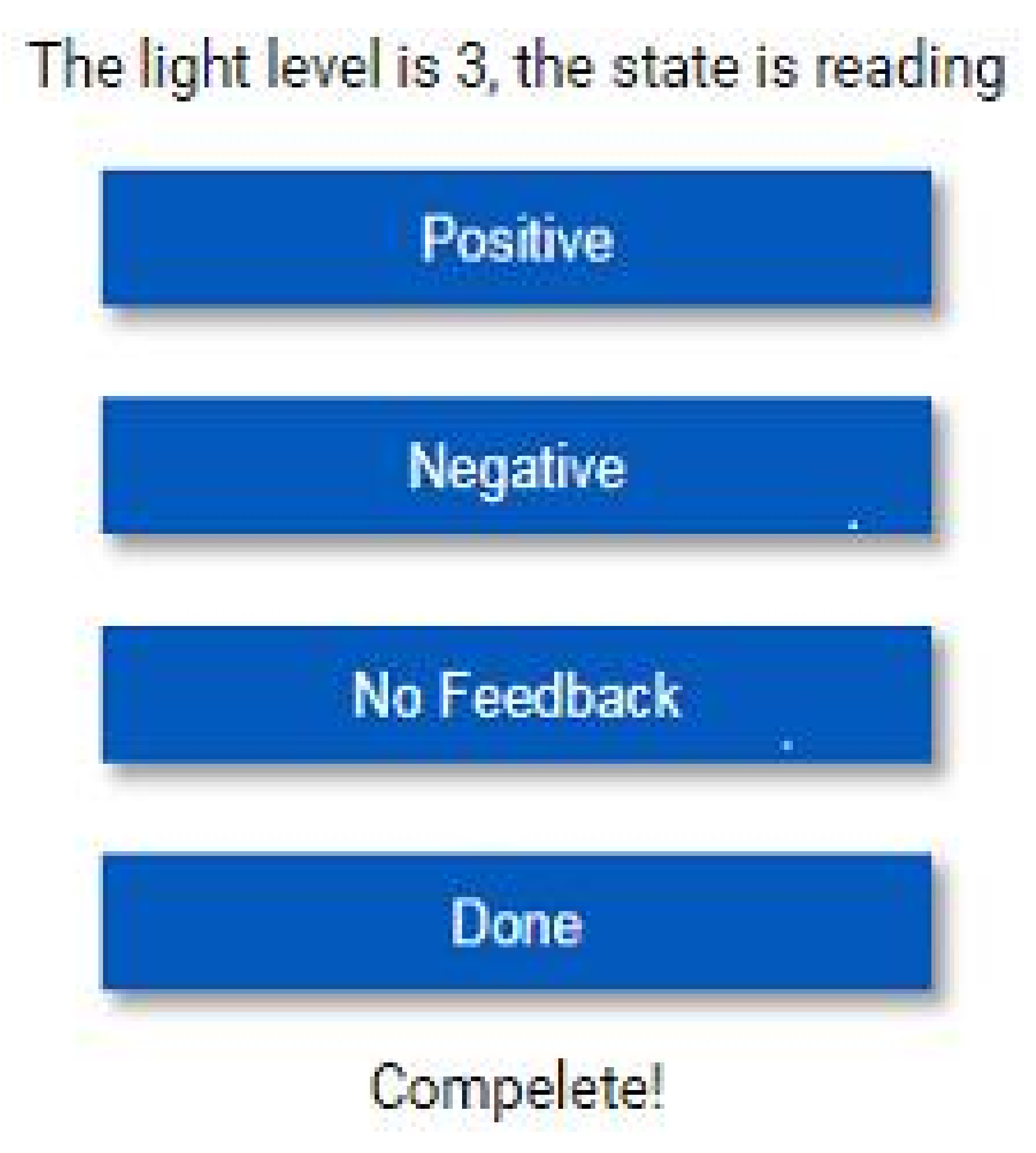}\label{gui_feedback}
}
\setlength{\abovecaptionskip}{0pt}
\caption{GUI designed for human experiment. (a) Users can change the light level directly by clicking the buttons. (b) Users can transfer their satisfaction by clicking the first three buttons.}\label{gui}
\end{figure}
We invited 40 students to participate in our experiment and built the experiment environment by a room with a computer and a YeeLight\footnote{\url{https://www.yeelight.com}} bulb that is able to be controlled remotely. The experiment is divided into four parts to evaluate four different algorithms (ABLUF, UCB, ISABL, QUERY) which are arranged randomly during the experiment. In each part, participants' preferred light levels in three states (reading, watching videos and talking with others) are learned and the order of these states is fixed. The number of optional light levels is 10 ($0\%, 11\%, 22\%,\dots, 99\%$ of the bulb's maximum lightness), which is a balance of two concerns. 1) It is difficult for human beings to detect the difference between two adjacent light levels when the number is larger. 2) A smaller number of light levels would potentially degrade the comfort of users. In the QUERY method, we directly ask users to adjust light levels rather than providing feedback. 

Users can interact with our system by GUIs shown in Figure \ref{gui}.
The current light level and state are shown on the top of the web site and the `Done' button is used to transform the current state to the next one.
For QUERY method, users can directly click the buttons with numbers to change the light level in Figure \ref{gui_query}. A larger number corresponds to a brighter light level. 
For other methods, we design a similar GUI shown in Figure \ref{gui_feedback}. 
The first three buttons are designed for collecting different kinds of feedback. To collect feedback accurately, the participants can press the `No Feedback' button to provide `no feedback'.
\begin{table}[!t]
\setlength{\belowcaptionskip}{0.25cm}
\caption{The result of human experiment ($mean \pm std. \ deviation$) under two metrics. The performance of our algorithm is statistically superior to others.}
    \centering
    \scalebox{1.1}{
    \begin{tabular}{c|c|c}
    \hline
    \hline
    Algorithm & distance/state & \#steps/state\\ 
    \hline
    UCB & $40.48 \pm 11.09$ & $14.72 \pm 3.49$ \\
    \hline
     ISABL & $23.22 \pm 11.82$ &  $8.21 \pm 3.37$\\
    \hline
     QUERY &$17.75 \pm 4.68$ & $7.85 \pm 2.32$\\
    \hline
     ABLUF & $\mathbf{13.33 \pm 4.59}$ & $\mathbf{5.96 \pm 1.50}$\\
    \hline
    \hline
    \end{tabular}}
    \label{human_result}
\end{table}

\textbf{When to stop?} For these methods, the algorithm chooses a light level from 10 candidates according to the participant's feedback. Then, participants are asked to provide feedback. The maximum number of steps and the time for each algorithm are not constrained. After participants think that the system has learned their preferences, they are also asked to stay in the current state for 1 minute and then interact with two more steps. If the participant does not want to change the light level and the algorithm maintains his/her favorite light level when the participant interacts with the agent in the two additional steps, we think that the algorithm has converged and learned participants' preference. Then, participants can click the `Done' button and the state will be transferred to the next one until all of the states are traversed. 
For the QUERY method, participants can explore and select their favorite light levels in each state without any constraint. In each state, they can choose other light levels through the GUI if they feel uncomfortable. After they think a light level is comfortable, we ask them to stay in the current lighting situation for 1 minute to ensure that their feeling is stable. If their feeling does not change, they can click `done' button to transfer to the next state. Otherwise, they could continue to choose another light level.

\begin{figure}[!t]
\setlength{\belowcaptionskip}{-0.5cm}
\setlength{\abovecaptionskip}{0pt}
\centering
\subfigure[Distance/state]{
\includegraphics[width=0.22\textwidth]{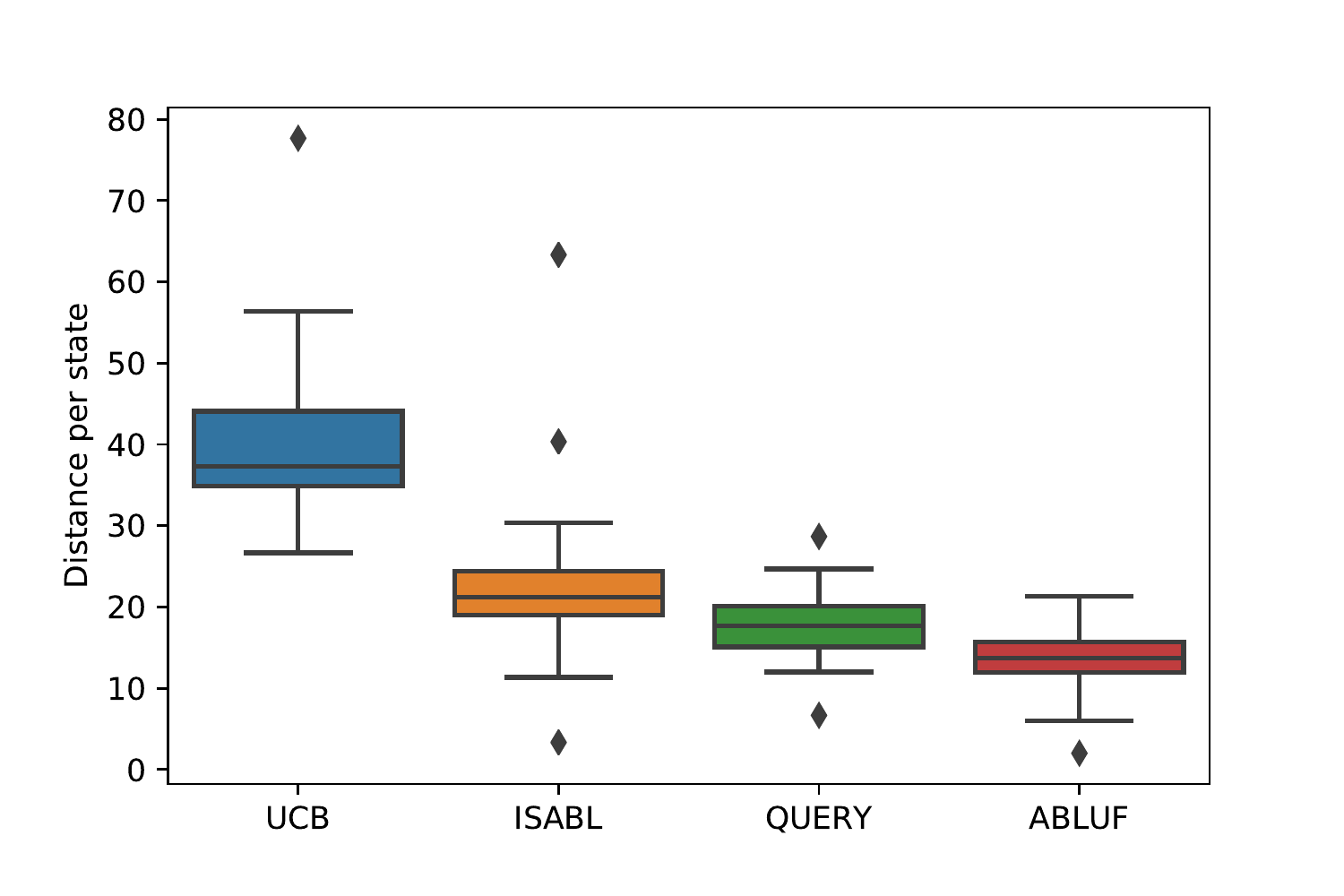}
}
\subfigure[\#steps/state]{
\includegraphics[width=0.22\textwidth]{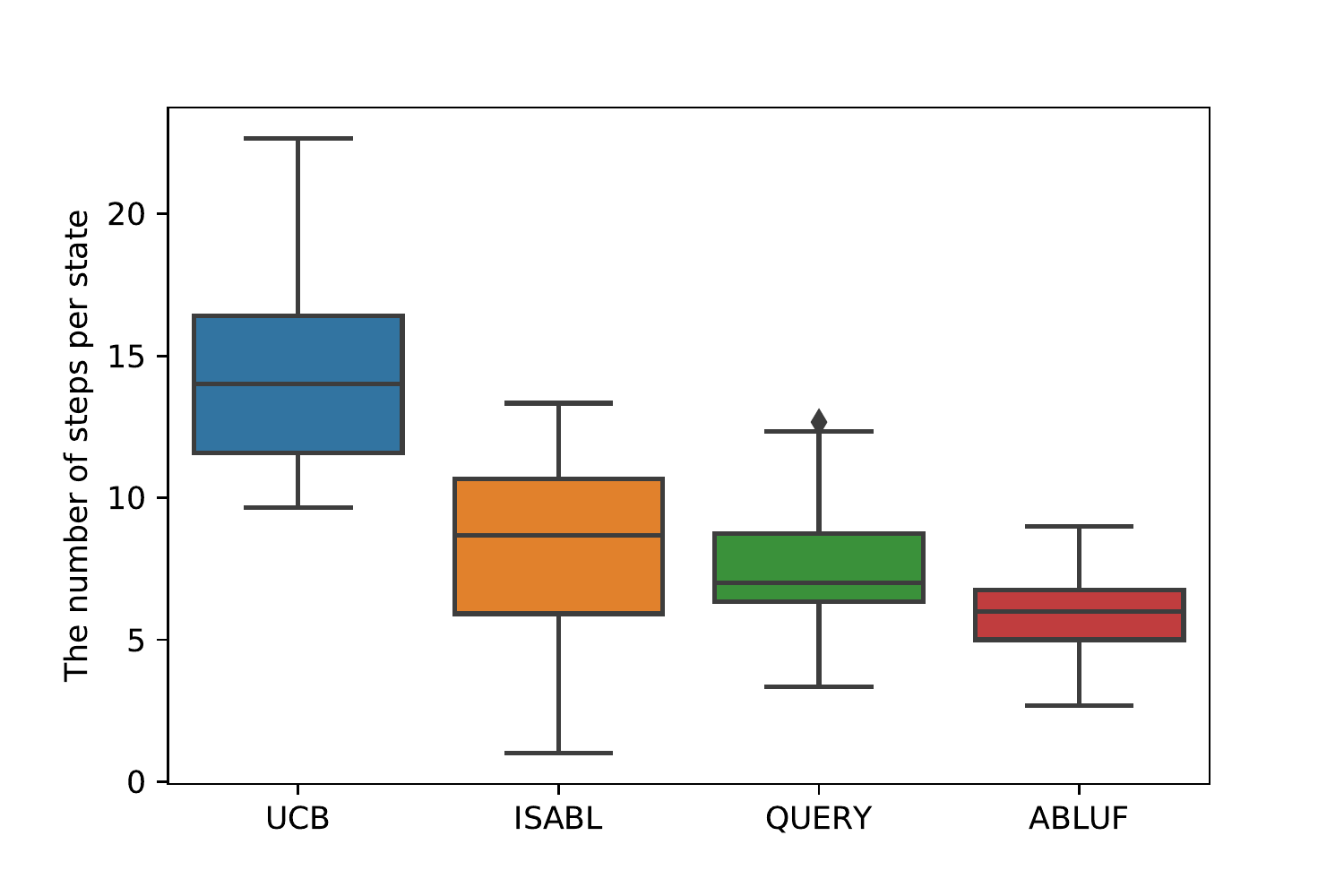}
}
\caption{The result of human experiment under accumulative distance and steps.}\label{human}
\end{figure}
\textbf{Metrics.} For this experiment, the step required to finish the training as the metric could be used as the metric. The other two metrics introduced in the last section is not applicable since the setting of this experiment is different. Since users only end the experiment when they feel comfortable, the learned policy is the optimal one. And it is difficult to know the probability that a user feels comfortable when the light level is not his preferred one. Alternatively, we use accumulative distance $d = \sum_{t=0}^T [a_t-\lambda(s)]^2$ between the current action and the optimal action as a metric to evaluate the degree of discomfort. Intuitively, if the distance is large, the user are not likely to feel comfortable. In practice, if one light level is selected in the last few steps repeatedly, we will ignore these steps since what we are concerned about is the total number of steps required for learning rather than verifying. 

\textbf{Results.} The result of our human experiment is shown in Figure \ref{human} and Table \ref{human_result}. The ABLUF method is statistically superior to the comparing algorithms using Wilcoxon two-sided signed-rank test ($p < 0.01$). In the QUERY method, participants cannot select their favorite light levels with a small number of steps, since they need to compare similar light levels to find their favorite ones and their strategies of finding preferred light levels are radical. For example, when the light level is too bright, users are prone to select a very dark one and then use more steps to find the optimal one.
This observation is supported by \cite{boutilier2002pomdp,french1986decision} which show that utility functions are very difficult for users to access, which indicates that they cannot tell their preferred light levels accurately without comparison. 
Our algorithm designs a more efficient way to explore various light levels according to participants' feedback.
\begin{figure}[!t]
\setlength{\belowcaptionskip}{-0.4cm}
\centering
\subfigure[Rats caught per step \newline when $|s|=4$]{
\includegraphics[width=0.225\textwidth]{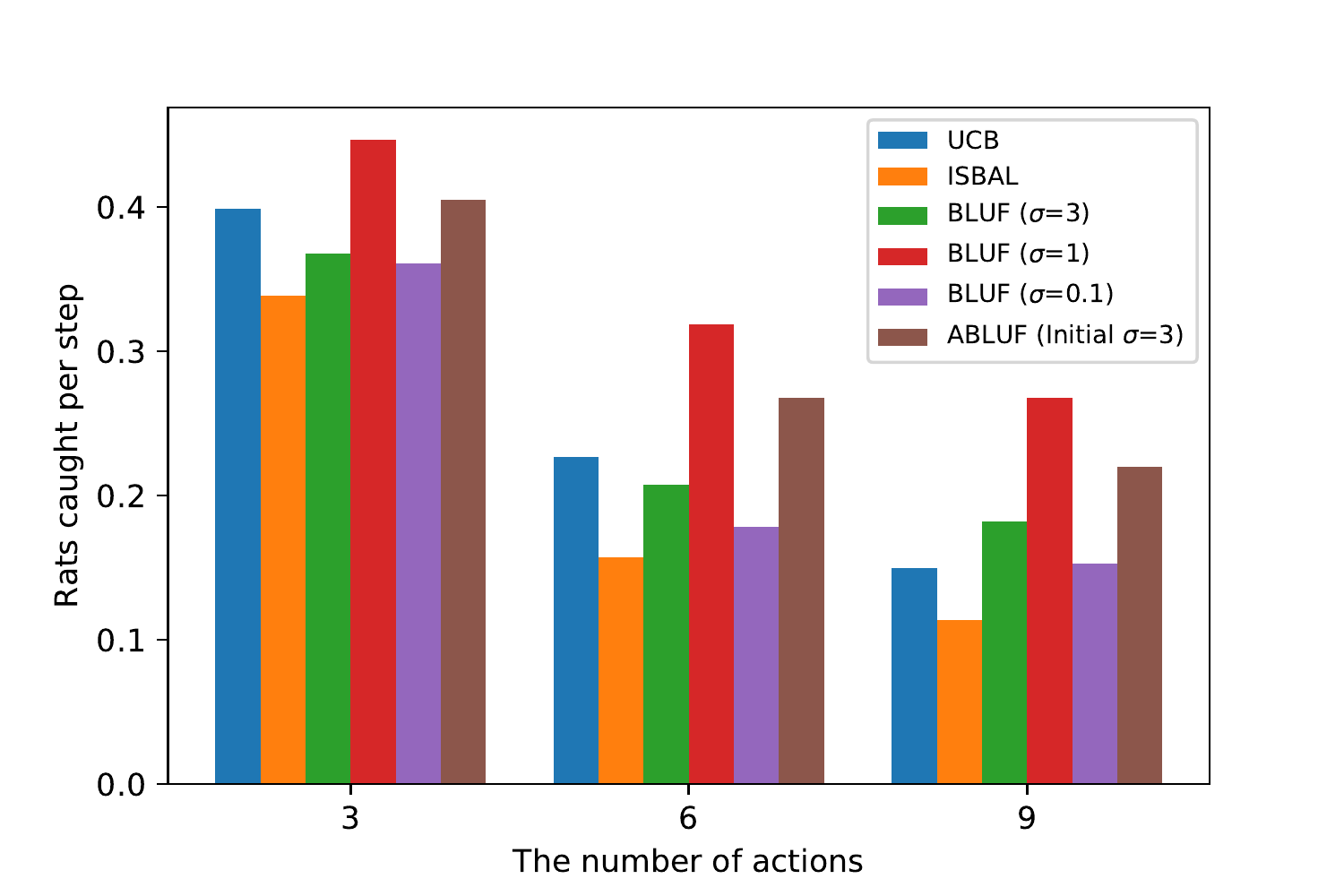}
}
\subfigure[The gap between policies when $|s|=4$]{
\includegraphics[width=0.225\textwidth]{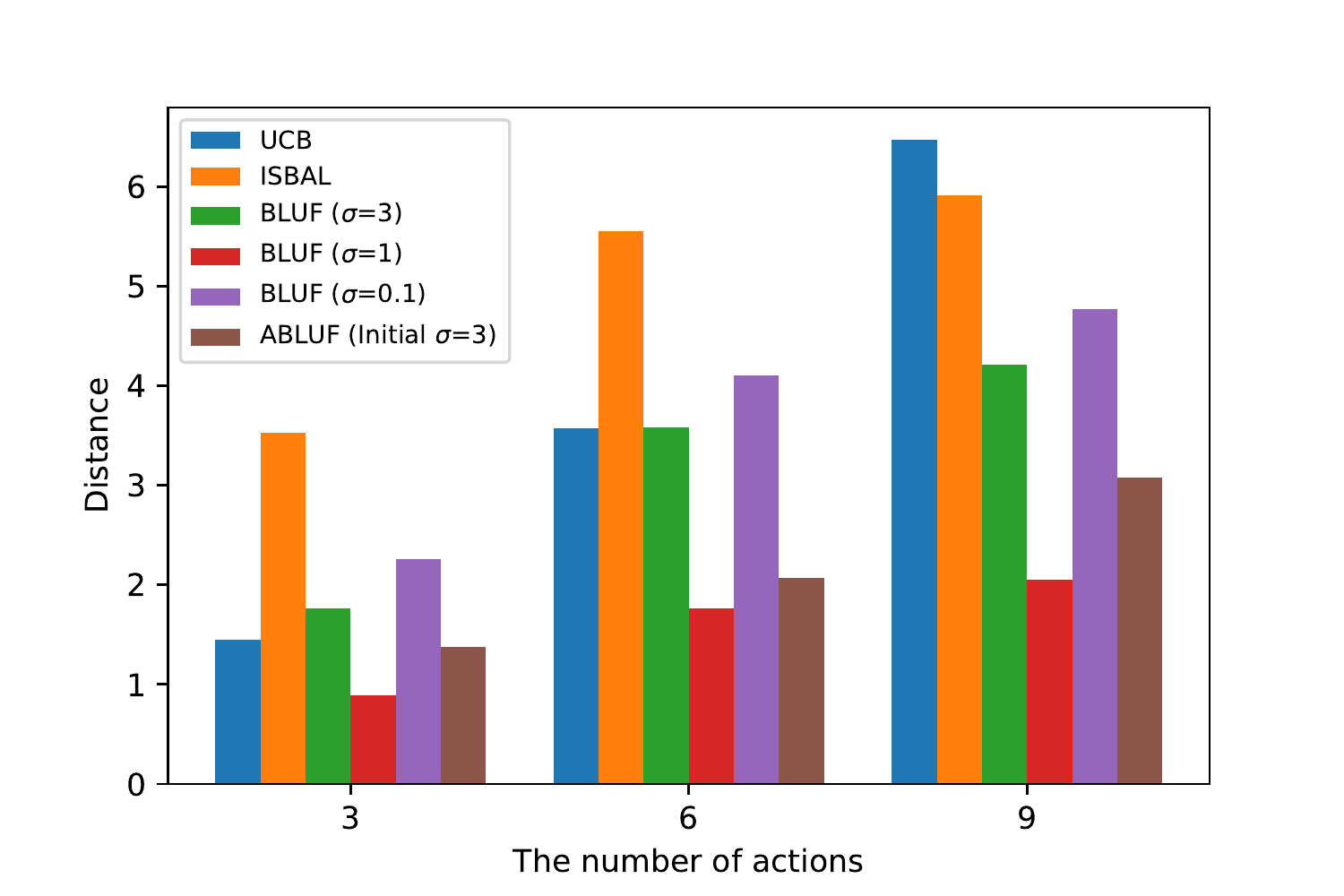}
}\\
\subfigure[Rats caught per step \newline when $|a|=3$]{
\includegraphics[width=0.225\textwidth]{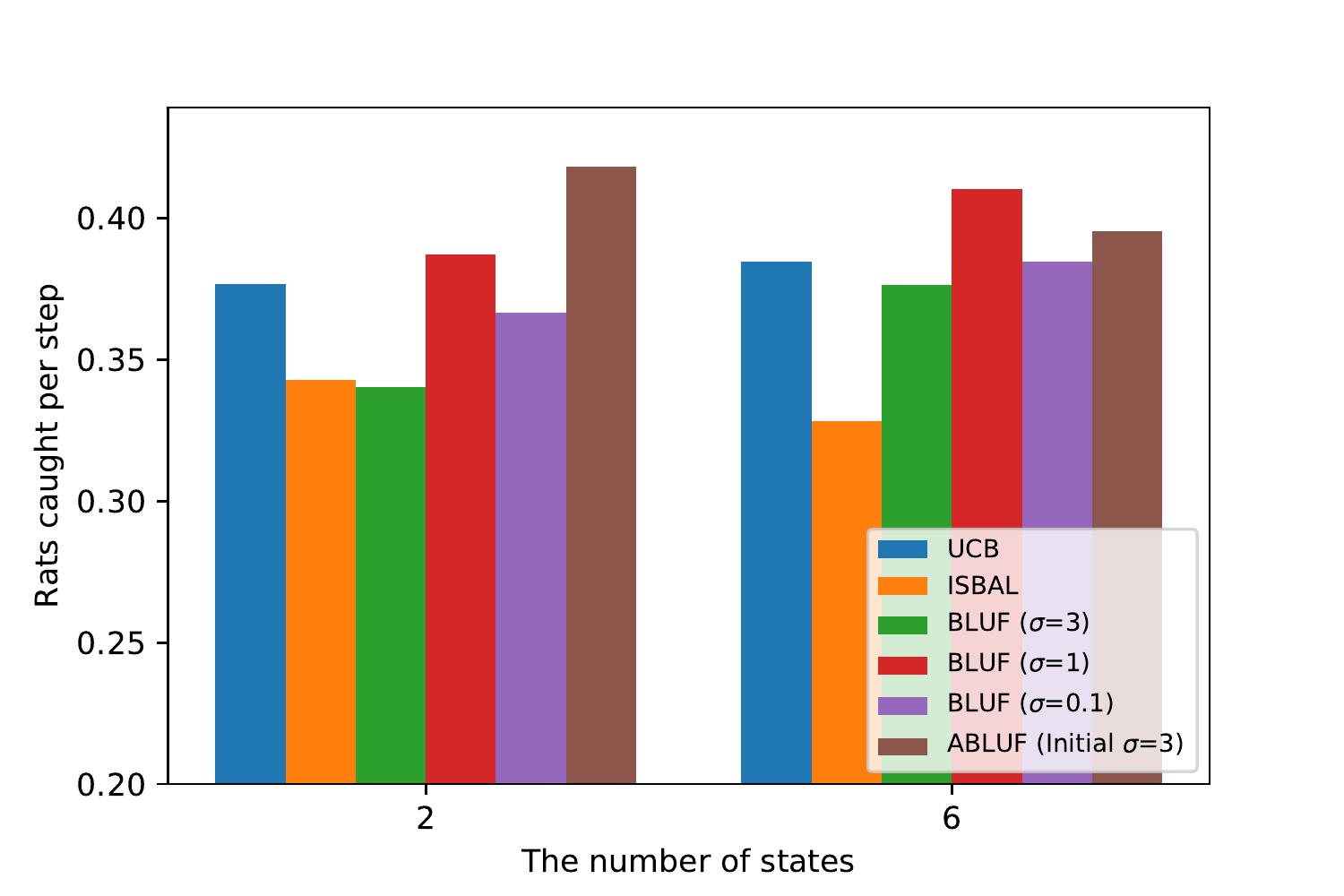}
}
\subfigure[The gap between policies when $|a|=3$]{
\includegraphics[width=0.225\textwidth]{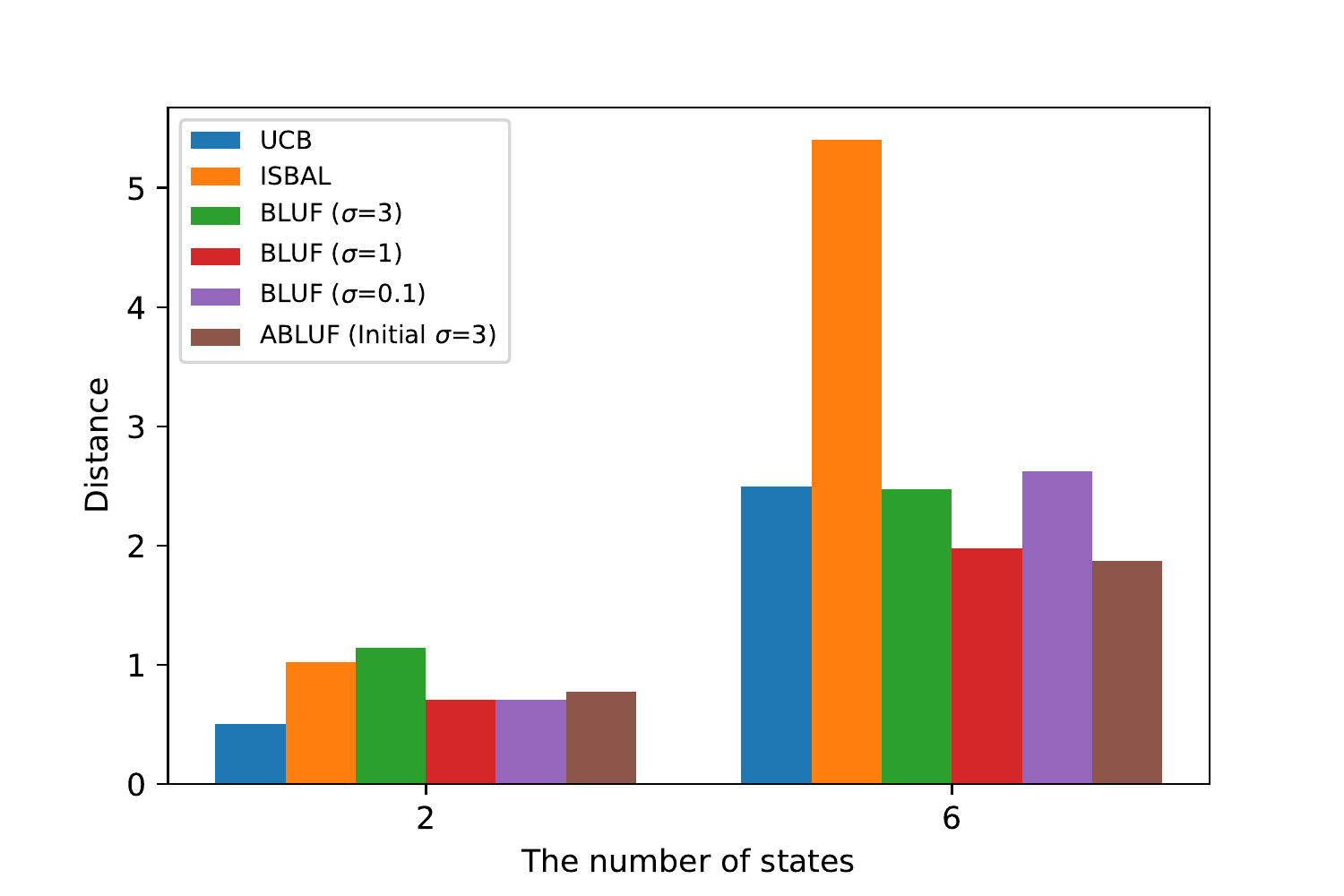}
}
\setlength{\abovecaptionskip}{0pt}
\caption{The result of dog training experiment with different numbers of actions and states.}\label{dog_s}
\end{figure}

\subsection{Performance in Various Synthetic Environmental Settings}
In this subsection, we implement experiments with simulated trainers/users in the above mentioned environments to answer three questions: 1) How does the number of actions and states influence the performance of different algorithms. 2) Can our algorithm learn the true value of $\sigma$ precisely? 3) Can our algorithm perform well when trainers does not follow our feedback model?
The experimental results show that the ABLUF algorithm is better than two baselines significantly ($p<0.01$ by t-test) and can learn the value of $\sigma$ accurately in most of cases.

\textbf{Ablation.} To illustrate the importance of updating the feedback model, we also compare with Bayesian Learning for Uncertain Feedback (BLUF) algorithm, a simplified version of our ABLUF algorithm, where we treat the parameter $\sigma$ as a constant input ($\sigma \in \{0.1, 1, 3\}$ in our experiments). This means that we have prior knowledge about the changed ratio of the probabilities at action $a$ given the distance $d(a,a^*_s)$. However, it is hard to get plenty of data to obtain such prior knowledge in the real world for each trainer. Thus, the BLUF algorithm is not feasible in our human experiment. We only compare with it to illustrate the impact of an accurate $\sigma$ in this simulated experiment.
\begin{figure}[!t]
\centering
\subfigure[Fixed \#states: 4]{
\includegraphics[width=0.22\textwidth]{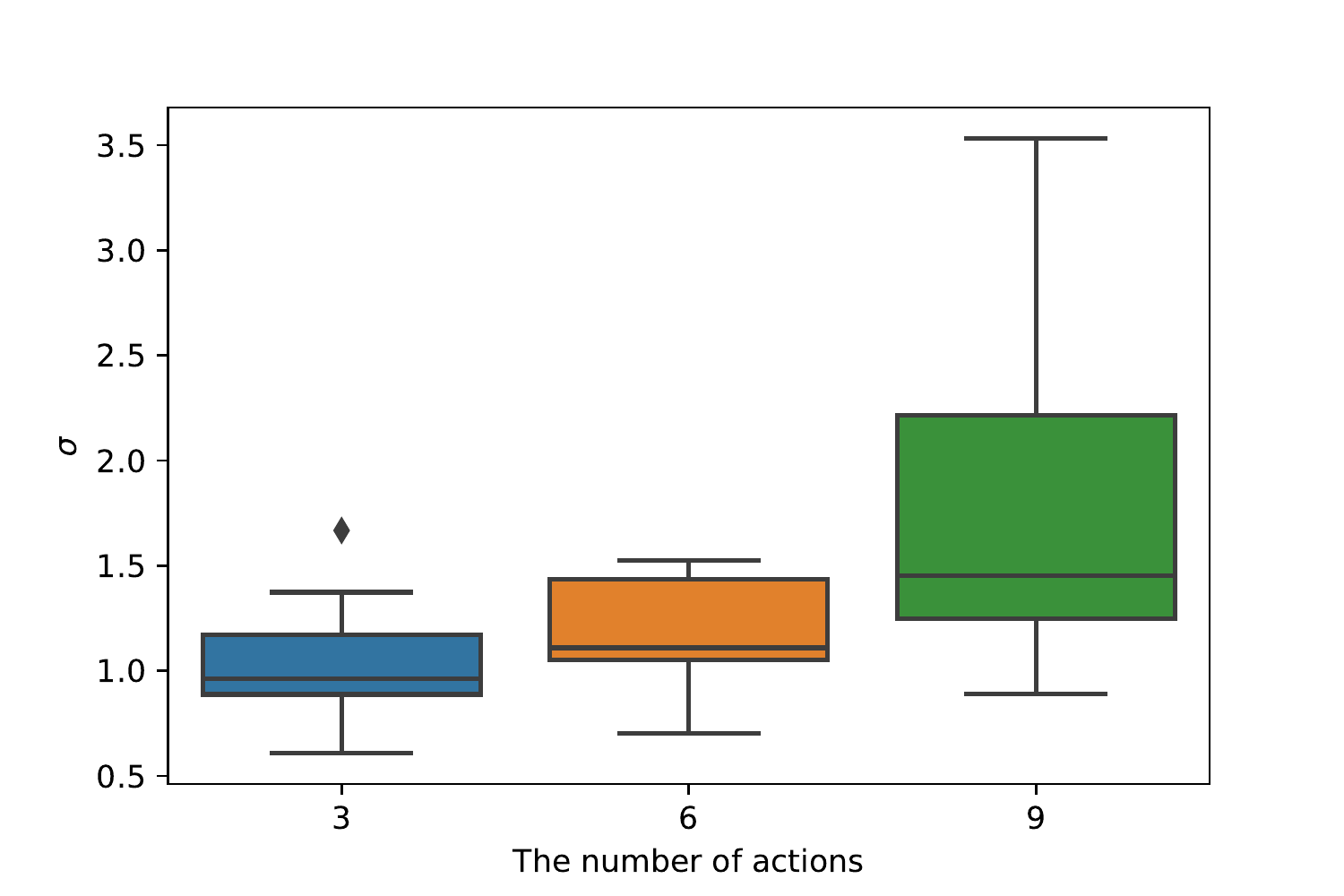}
}
\subfigure[Fixed \#actions: 3]{
\includegraphics[width=0.22\textwidth]{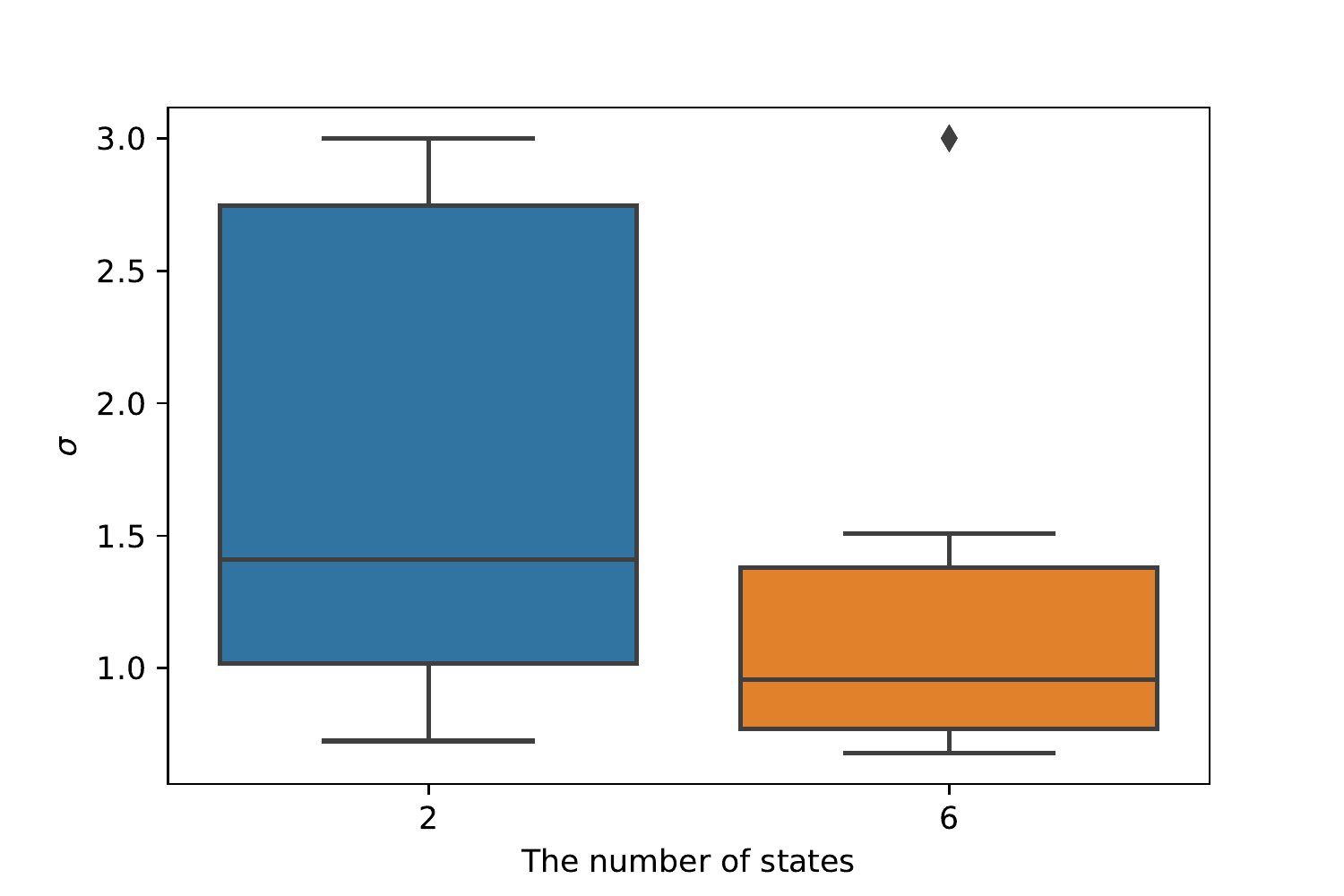}
}
\setlength{\abovecaptionskip}{0pt}
\caption{The values of $\sigma$ after training.}
\label{dog_sigma}
\end{figure}
\textbf{Answer for the first question.} 
For the synthetic data, the probabilities of obtaining different kinds of feedback follow our feedback model in which $\sigma = 1$ and $\mu$ is randomly generated. The optimal actions are generated randomly for various states. For the both environments, we set the number of actions to 3, 6 and 9 when the number of states is 4. When the number of actions is 3, we vary the number of states to 2 and 6. Trainers are allowed to interact with algorithms for 15 steps per state, which is similar to our human experiment.

For the dog training environment, Figure \ref{dog_s} shows the performance of all the  algorithms in different settings. The sub figures indicate that the change of the number of actions affects performance more significantly due to two reasons: 1) There are more sub-optimal actions when the number of actions is large, which increases the difficulty to figure out the optimal actions. 2) The worst case for two metrics becomes worse. The maximum distance of the optimal action and another action increases from $\sqrt{2}$ to $\sqrt{8}$ when the number of actions varies from 3 to 9. Similarly, the minimal probability of catching a rat declines from $e^{-2}$ to $e^{-32}$. Therefore, performance of all algorithms descends. Our ABLUF method outperforms all the baselines except the BLUF algorithm with accurate $\sigma=1$ known in advance. For other ablation methods with biased $\sigma$, our adaptive method achieves better performance, which shows that an accurate $\sigma$ is crucial.

However, the variation of states does not impact the number of the rats caught per step, since the probabilities of catching rats are similar for both the optimal action and a sub-optimal one. This conclusion is verified by the last figure, where the gap increases with respect to the number of states. The figure illustrates that the learned policy is worse when $|s|=6$. However, the rat caught per step does not decrease, which indicates that a sub-optimal policy does not affect much in this setting.
The performances of ABLUF and BLUF are similar when $|s|=2$. The reason would be that a simple setting could be solved by an inaccurate model such as ISABL.

\begin{figure}[!t]
\centering
\subfigure[\#states:4, \#actions: 3\label{a3}]{
\begin{minipage}[b]{0.22\textwidth}
\centering
\includegraphics[width=1\textwidth]{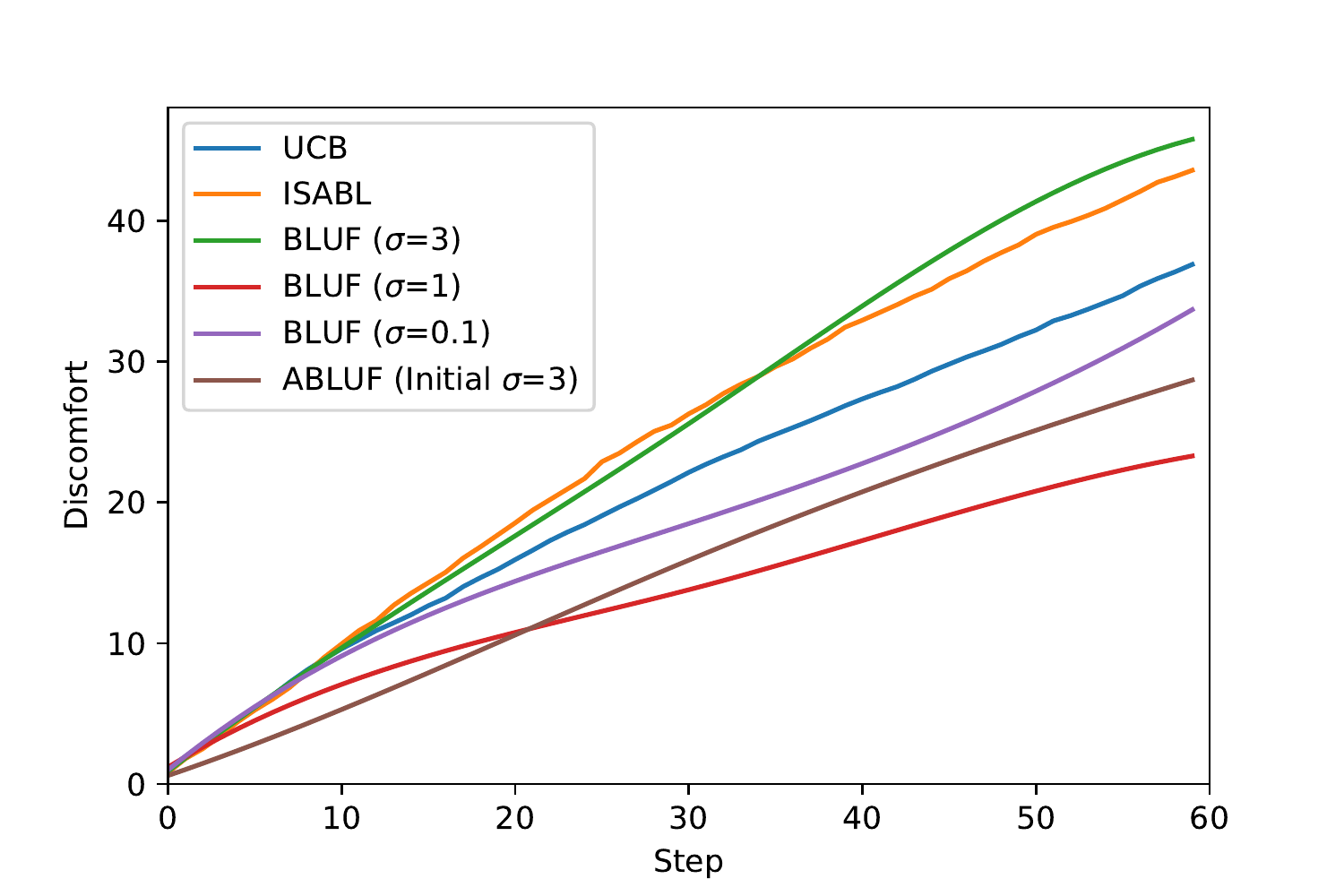}
\end{minipage}}
\subfigure[\#states:4, \#actions: 6]{
\begin{minipage}[b]{0.22\textwidth}
\centering
\includegraphics[width=1\textwidth]{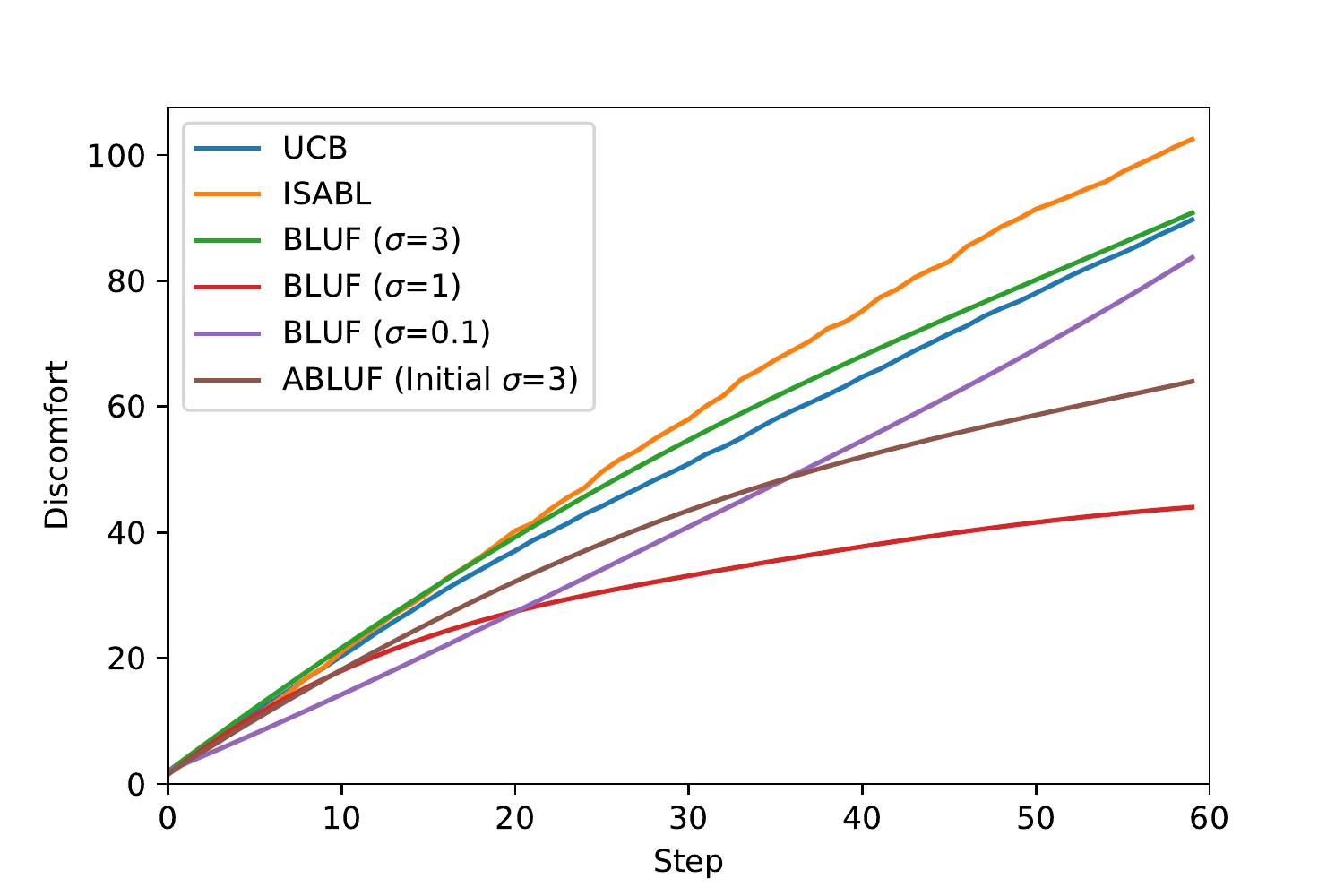}
\end{minipage}}\\
\subfigure[\#states:4, \#actions: 9\label{a9}]{
\begin{minipage}[b]{0.22\textwidth}
\centering
\includegraphics[width=1\textwidth]{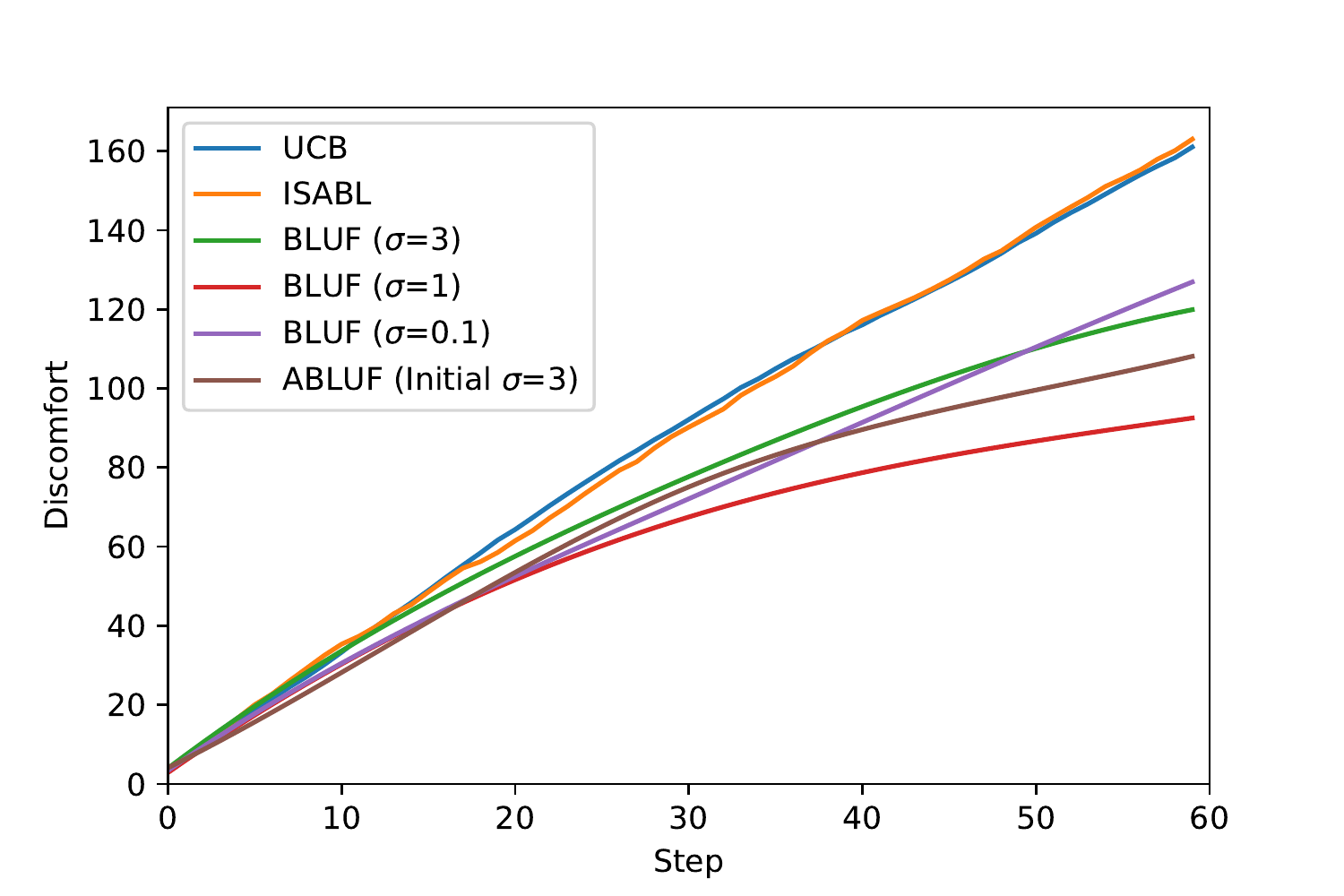}
\end{minipage}} 
\subfigure[\#states:2, \#actions: 3\label{s2}]{
\begin{minipage}[b]{0.22\textwidth}
\centering
\includegraphics[width=1\textwidth]{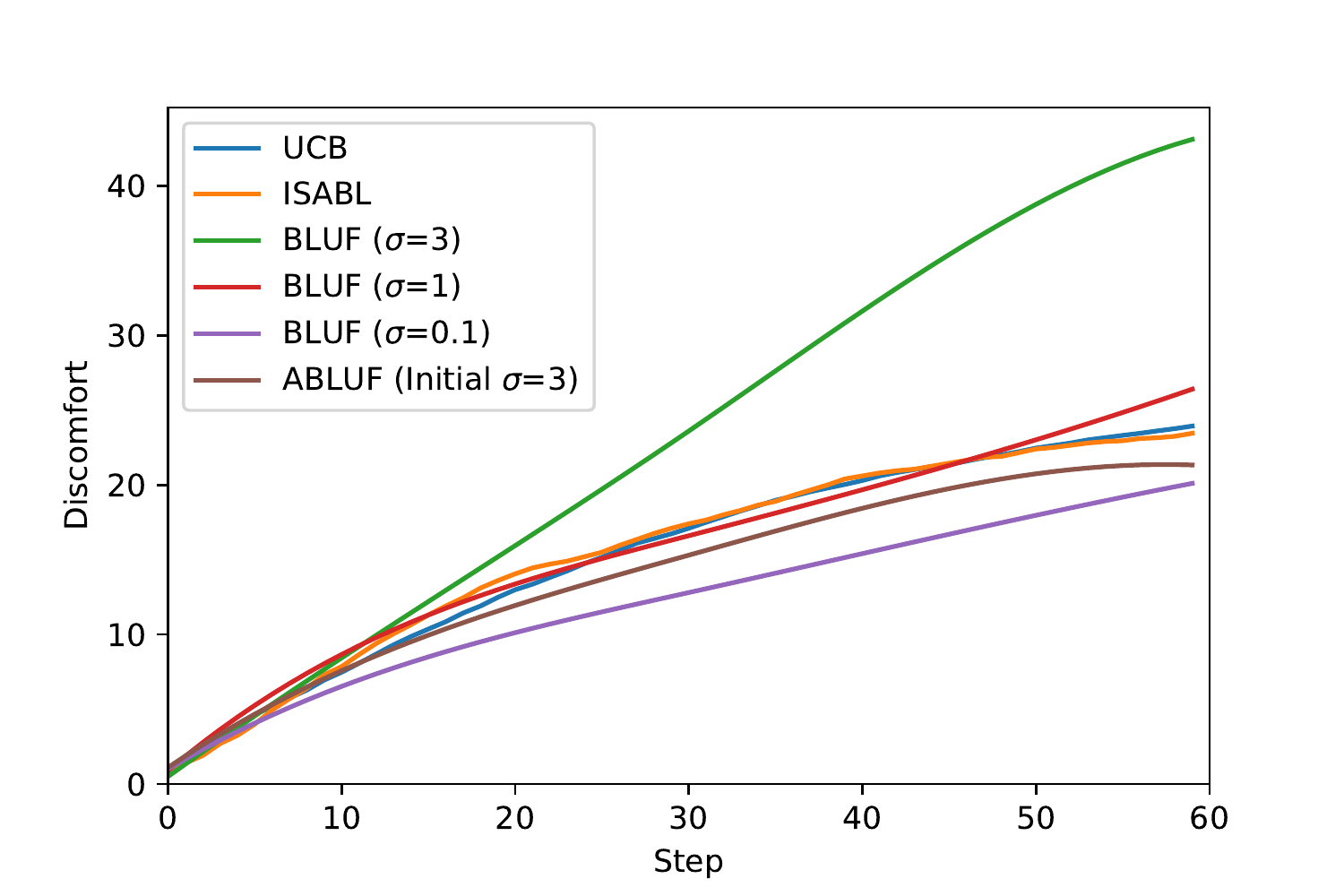}
\end{minipage}}\\
\subfigure[\#states:6, \#actions: 3\label{s6}]{
\begin{minipage}[b]{0.22\textwidth}
\centering
\includegraphics[width=1\textwidth]{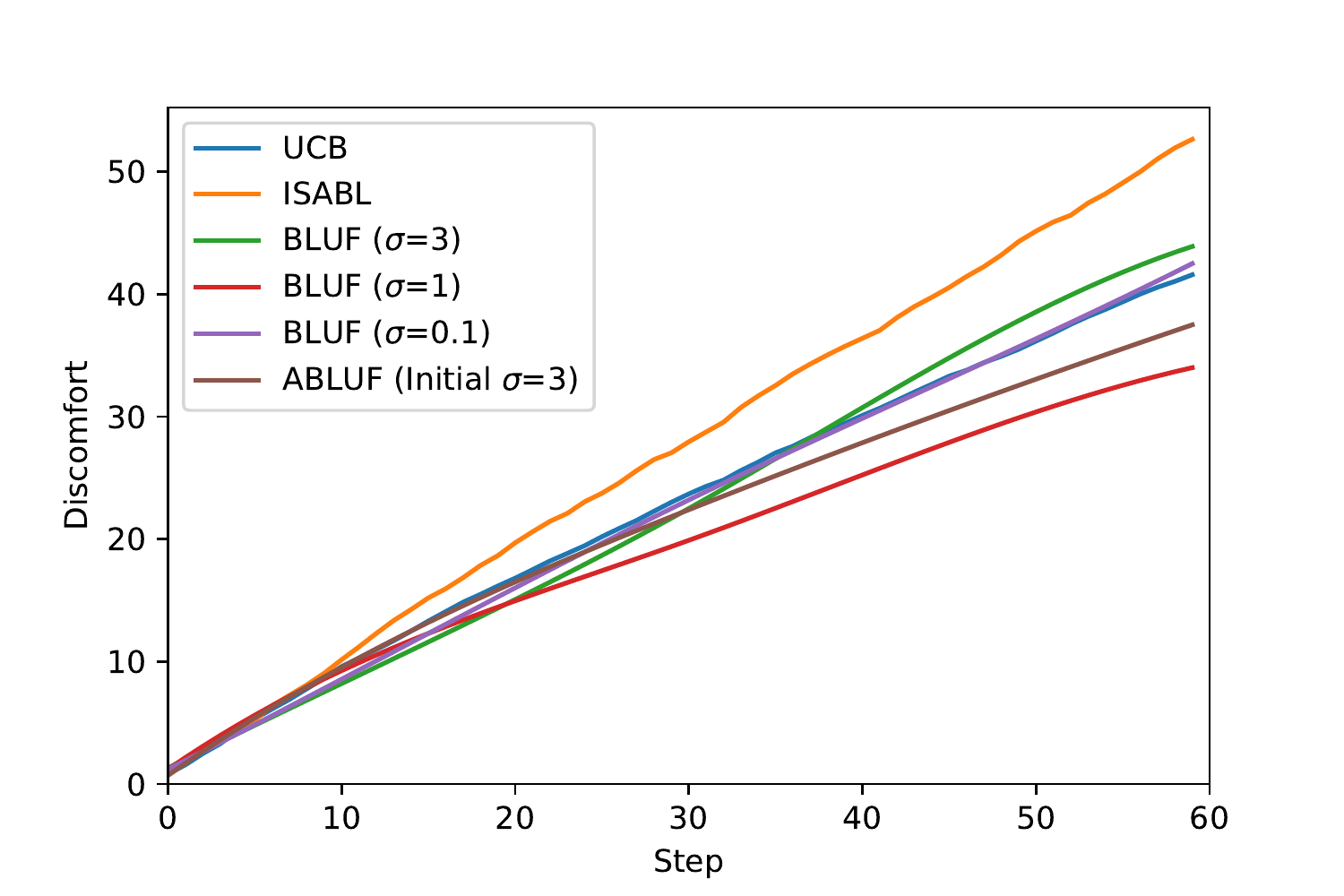}
\end{minipage}}
\subfigure[Randomly generated User (\#states:4, \#actions: 6)\label{ran}]{
\begin{minipage}[b]{0.22\textwidth}
\centering
\includegraphics[width=1\textwidth]{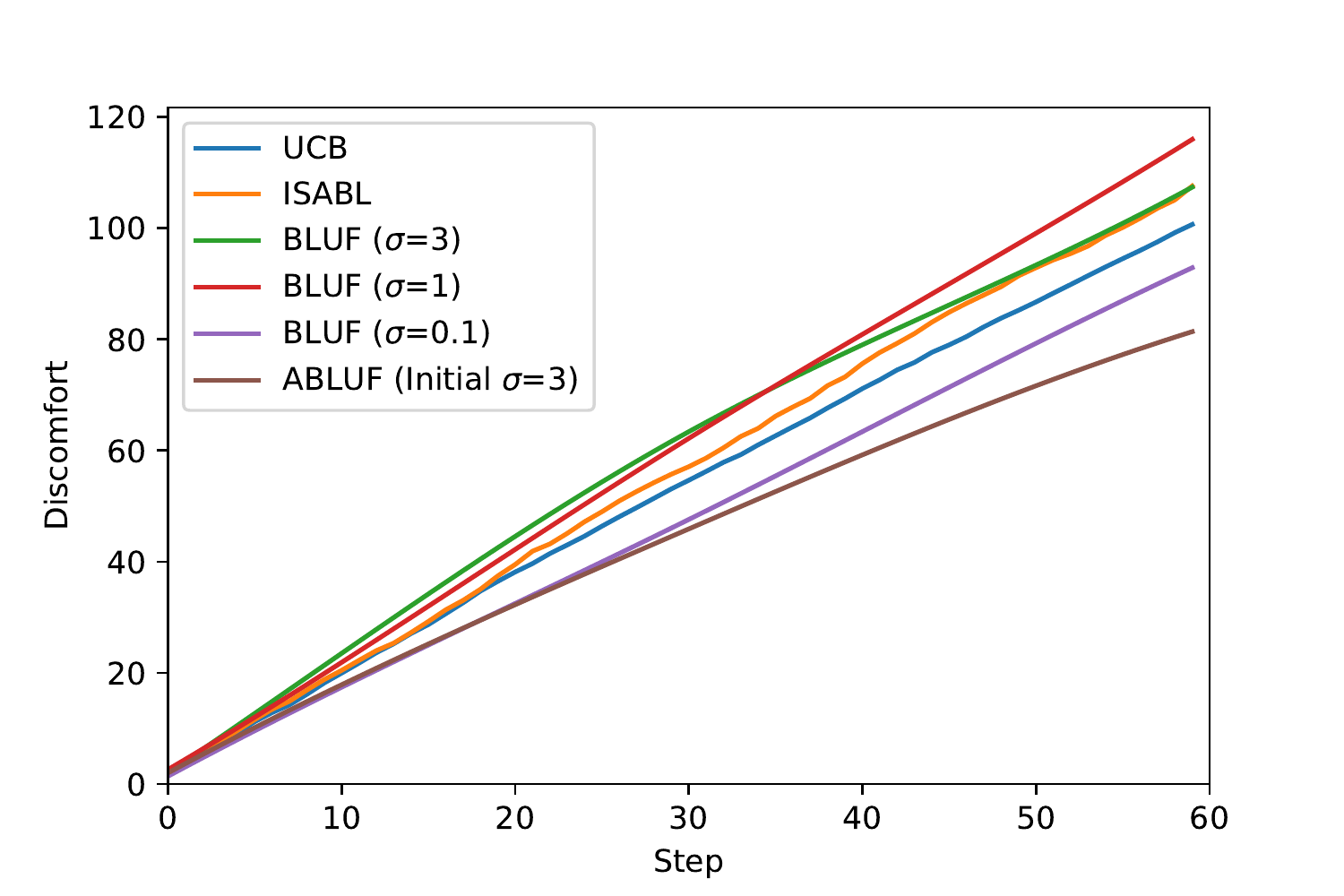}
\end{minipage}}
\setlength{\abovecaptionskip}{0pt}
\caption{Accumulative distance trend with the change of the number of states and actions.}
\label{action}
\end{figure}

For the lighting control environment, similar results are displayed in the Figures \ref{a3} to \ref{s6}. Since the maximum value of $[a-\lambda(s)]^2$ increases dramatically when we change the number of actions form 3 to 9. The accumulative distances of all the algorithms vary rapidly form Figure \ref{a3} to Figure \ref{a9}, while the increase is relatively small when the number of states becomes larger, which is the same as the dog training experiment.

\textbf{Answer for the second question.} Figure \ref{dog_sigma} illustrates the learning results of the parameter $\sigma$ summarised from two environments. The worst case appears when $|s|=4,|a| = 9$ and $|s|=2,|a| = 3$. In the simplest setting, the accuracy of $\sigma$ does not have a significant influence on convergence and the ABLUF algorithm figures out the optimal actions quickly without collecting enough feedback. Thus, $\sigma$ cannot be learned precisely. When $|s|=4$ and $|a| = 9$, since the limitation of steps is 15 per state, it is not enough for our algorithm converges to the optimal policy and thus the estimate of the ratio $ratio_a(f)$ is biased leading to the update towards wrong directions. This phenomenon would be eliminated with more steps and feedback, since for other settings, our algorithm can learn a relatively precise $\sigma$ with little variance.
\begin{figure}[!t]
\setlength{\belowcaptionskip}{-0.6cm}
\centering
\subfigure[Rats caught per step]{
\includegraphics[width=0.225\textwidth]{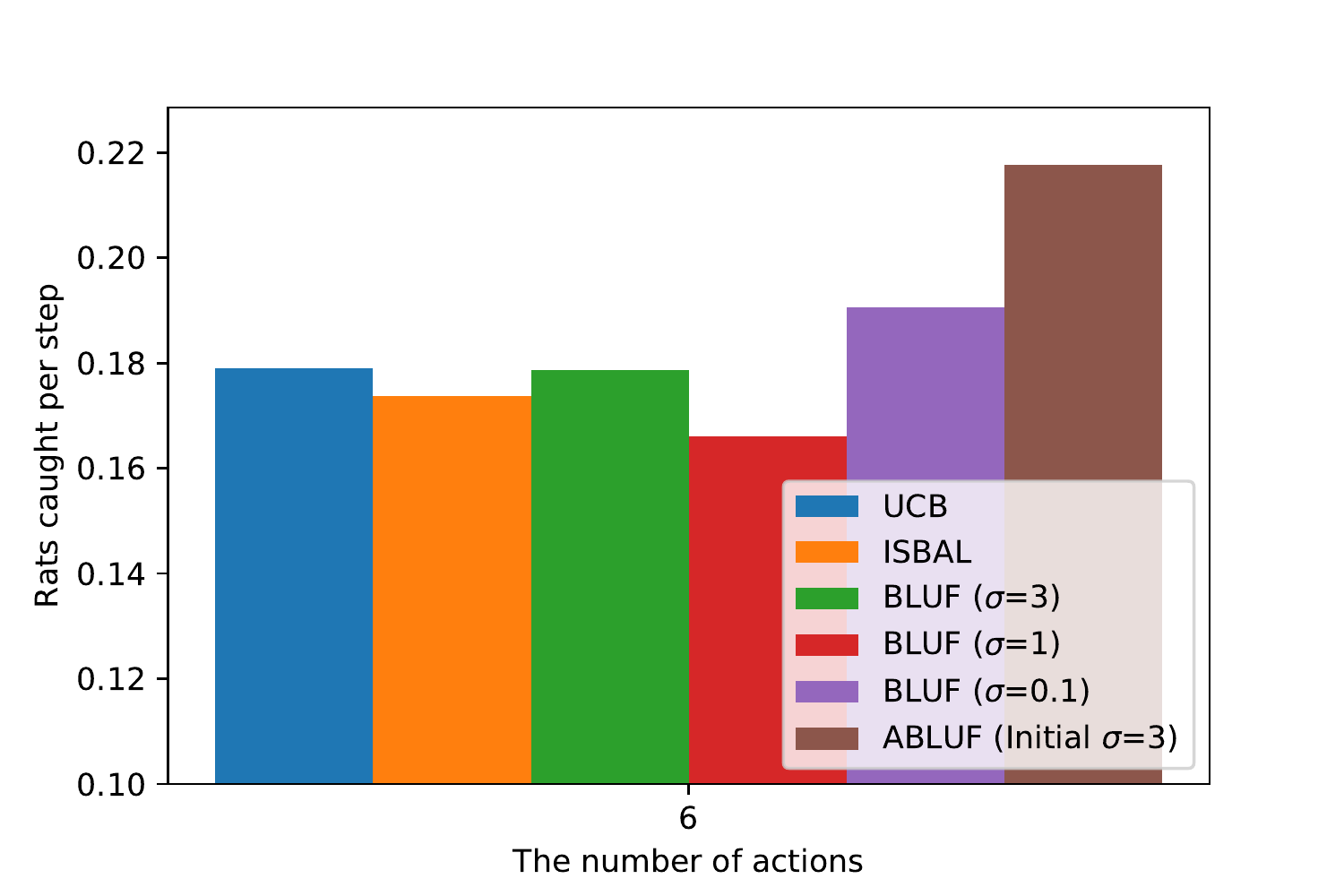}
}
\subfigure[The gap between policies]{
\includegraphics[width=0.22\textwidth]{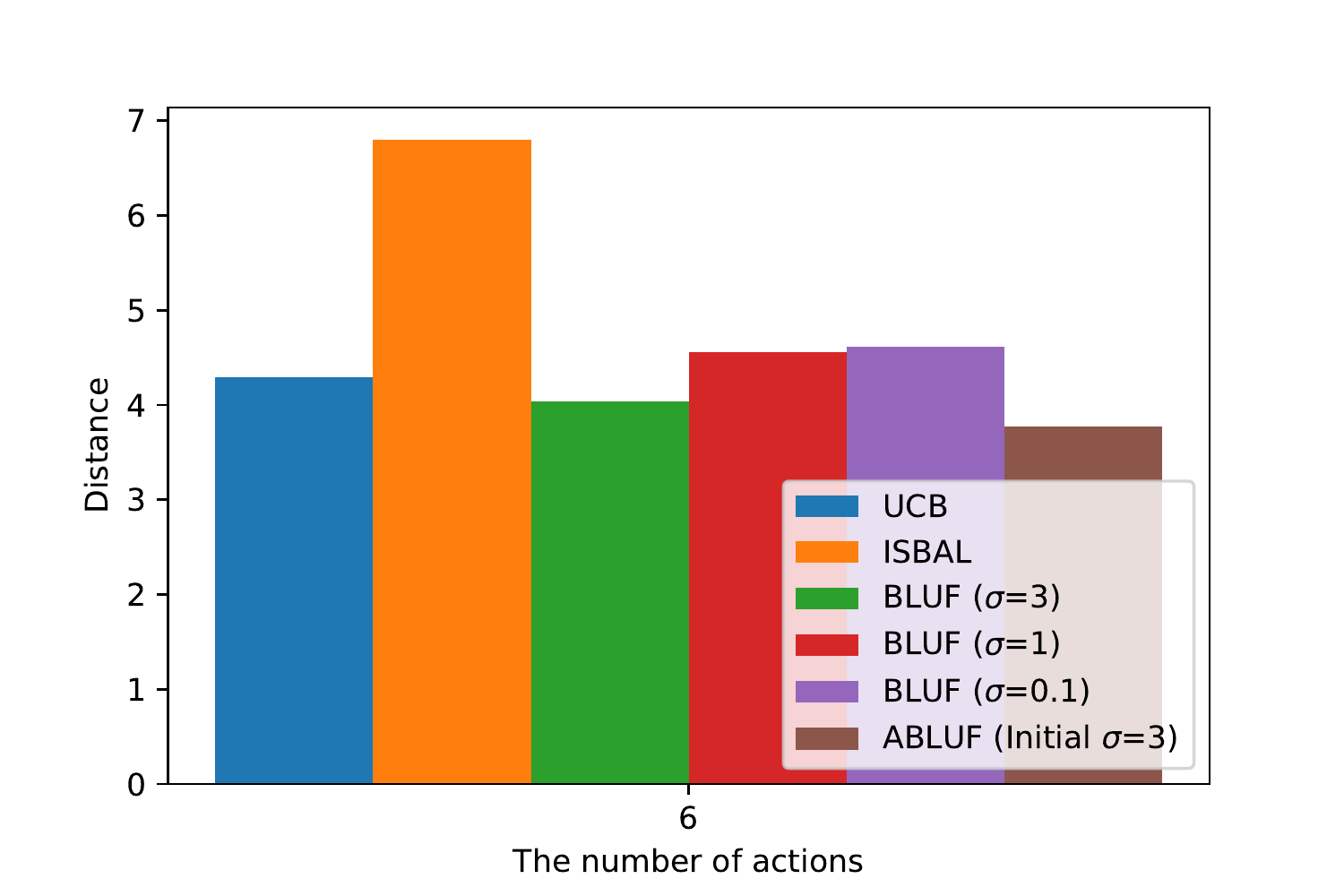}
}
\setlength{\abovecaptionskip}{0pt}
\caption{The result of dog training experiment with randomly generated trainers when $|s|=4$ and $|a|=6$.}\label{dog_random}
\end{figure}

\textbf{Answer for the third question (Robustness Analysis).} Additionally, we evaluate our algorithm in the worst case, in which the trainer does not follow our feedback model and the probabilities of providing different kinds of feedback are generated randomly. We set $|s|=4$ and $|a|=6$, which is the same as the human experiment for training virtual dogs.
We randomly generate simulated trainers and the only constraint is that the optimal action at each state has the highest probability of receiving positive feedback and the lowest probability of receiving negative feedback. The restriction is mild because of the definition of `the optimal action'. Figure \ref{dog_random} and Figure \ref{ran} demonstrate that even for the worst case, our algorithm outperforms the other two baselines. The ABLUF method performs the best in this case where we do not have any prior knowledge about human feedback models. Since we fix the value of $\sigma$ for BLUF algorithm, it performs not well when the true value of $\sigma$ differs from the prior knowledge given to BLUF. However, the ABLUF method can learn an approximated model even if these trainers do not follow our feedback model and thus is robust.
\section{Conclusion}
In this paper, we consider the uncertainty when humans provide feedback. More specifically, trainers are likely to provide positive feedback, negative feedback and no feedback to any action no matter if the action is optimal or not. To address this issue, we propose a novel feedback model that has two sets of parameters to control the shape of functions in the model. An approximated Expectation Maximization (EM) algorithm combined with Gradient Descent (GD) method is proposed to learn an optimal policy and update the feedback model simultaneously. To illustrate the performance of the novel method, we implement experiments on both synthetic scenarios and two different real-world scenarios with human participants. Experimental results indicate that our algorithm outperforms baselines under multiple metrics. Moreover, robustness analysis shows that our algorithm performs well even if trainers do not follow our feedback model.

\section{Acknowledgements}
This work was supported by the Delta Electronics Inc., National Research Foundation of Singapore and Nanyang Technological University.

\bibliographystyle{plain}
\bibliography{reference}
\end{document}